\documentclass[twoside,11pt]{article}
\usepackage{jmlr2e}
\usepackage[utf8]{inputenc} % allow utf-8 input
\usepackage[T1]{fontenc}    % use 8-bit T1 fonts
\usepackage{booktabs}       % professional-quality tables
\usepackage{nicefrac}       % compact symbols for 1/2, etc.
\usepackage{microtype}      % microtypography

\usepackage{bbm}

\usepackage[margin=1in]{geometry}
\usepackage{verbatim,float,url,dsfont}
\usepackage{graphicx,subcaption,psfrag}
\usepackage{algorithm,algorithmic}
\usepackage{mathtools,enumitem}
\usepackage{multirow}

\makeatletter
\newcommand*\rel@kern[1]{\kern#1\dimexpr\macc@kerna}
\newcommand*\widebar[1]{%
  \begingroup
  \def\mathaccent##1##2{%
    \rel@kern{0.8}%
    \overline{\rel@kern{-0.8}\macc@nucleus\rel@kern{0.2}}%
    \rel@kern{-0.2}%
  }%
  \macc@depth\@ne
  \let\math@bgroup\@empty \let\math@egroup\macc@set@skewchar
  \mathsurround\z@ \frozen@everymath{\mathgroup\macc@group\relax}%
  \macc@set@skewchar\relax
  \let\mathaccentV\macc@nested@a
  \macc@nested@a\relax111{#1}%
  \endgroup
}
\makeatother

\newcommand{\argmin}{\mathop{\mathrm{argmin}}}

\newcommand{\minimize}{\mathop{\mathrm{minimize}}}
\newcommand{\st}{\mathop{\mathrm{subject\,\,to}}}

\def\R{\mathbb{R}}

\def\E{\mathbb{E}}
\def\P{\mathbb{P}}
\def\T{\mathsf{T}}

\def\th{^{\text{th}}}

\usepackage{mdwlist}
\usepackage{tikz}

\usepackage[all]{xy}
\CompileMatrices

\newcommand{\cbr}[1]{\left\{#1\right\}}

\newcommand{\Acal}{\mathcal{A}}

\newcommand{\Scal}{\mathcal{S}}
\newcommand{\Tcal}{\mathcal{T}}

\newcommand{\Pcal}{\mathcal{P}}

\newcommand{\Xcal}{\mathcal{X}}

\newcommand{\Ical}{\mathcal{I}}

\newcommand{\Gcal}{\mathcal{G}}

\DeclareMathOperator*{\range}{\mathrm{range}}

\newcommand{\intset}[1]{\cbr{1..n}}

\definecolor{firebrick}{rgb}{0.7, 0.13, 0.13}

\definecolor{skybrick}{rgb}{0.13, 0.13, 0.7}

\def \wtj {w^\tau_j}
\def \wtnj {w^{\tau,\nu}_j}

\def \pinball {\psi}
\def \nocrossx {\Xcal_0} % the set of x we use to enforce noncrossing
\def \wis {\mathrm{WIS}}

\def \iso {\mathbb{K}} % isotonic cone
\def \sort {\mathtt{Sort}} % sorting operator
\def \isoproj {\mathtt{IsoProj}} % isotonic projection operator
\def \softmax{\mathtt{SoftMax}} % soft max operator
\def \mms{\mathtt{MinMaxSweep}} % min-max sweep operator

\def\d{\mathsf{d}}
\usepackage{ragged2e}
\usepackage{subfiles}
\usepackage{xfrac}
\usepackage{diagbox}

\usepackage{lastpage}
\jmlrheading{24}{2023}{1-\pageref{LastPage}}{7/22; Revised
2/23}{4/23}{22-0799}{Rasool Fakoor, Taesup Kim, Jonas Mueller, Alexander J. Smola, Ryan J. Tibshirani}

\ShortHeadings{Flexible Model Aggregation for Quantile Regression}{Fakoor, Kim, Mueller, Smola, Tibshirani}
\firstpageno{1}

\graphicspath{{./final_figs/}}

\begin{document}
\title{Flexible Model Aggregation for Quantile Regression}

\author{\name Rasool Fakoor \email fakoor@amazon.com\\
  \addr Amazon Web Services 
  \AND
  \name Taesup Kim \email taesup.kim@snu.ac.kr\\
  \addr Seoul National University 
  \AND
  \name Jonas Mueller \email jonas@cleanlab.ai\\
  \addr Cleanlab 
  \AND
  \name Alexander J. Smola \email smola@amazon.com\\
  \addr Amazon Web Services  
  \AND
  \name Ryan J. Tibshirani \email ryantibs@cmu.edu\\
  \addr Amazon Web Services\\
  Carnegie Mellon University
}

\editor{Mladen Kolar}

\maketitle 
\begin{abstract}
Quantile regression is a fundamental problem in statistical learning motivated
by a need to quantify uncertainty in predictions, or to model a diverse
population without being overly reductive. For instance, epidemiological
forecasts, cost estimates, and revenue predictions all benefit from being able
to quantify the range of possible values accurately. As such, many models have
been developed for this problem over many years of research in statistics,
machine learning, and related fields.  Rather than proposing yet another (new)
algorithm for quantile regression we adopt a meta viewpoint: we investigate
methods for aggregating any number of conditional quantile models, in order to
improve accuracy and robustness. We consider weighted ensembles where weights
may vary over not only individual models, but also over quantile levels, and
feature values. All of the models we consider in this paper can be fit using
modern deep learning toolkits, and hence are widely accessible (from an
implementation point of view) and scalable.  To improve the accuracy of the
predicted quantiles (or equivalently, prediction intervals), we develop tools
for ensuring that quantiles remain monotonically ordered, and apply conformal
calibration methods. These can be used without any modification of the original
library of base models. We also review some basic theory surrounding quantile
aggregation and related scoring rules, and contribute a few new results to this
literature (for example, the fact that post sorting or post isotonic regression
can only improve the weighted interval score). Finally, we provide an extensive
suite of empirical comparisons across 34 data sets from two different benchmark
repositories. 
\end{abstract} 


\section{Introduction}

Consider the problem of assessing the height of a child. Common practice is to
consult a growth chart, such as the ones provided by the CDC
\citep{kuczmarski2000cdc} and to review the distribution of heights, as relevant
to the age and sex of the child. In doing so, the medical practitioner performs
quantile regression, conditioning their estimates on covariates (age, sex) to
obtain a conditional height distribution. While it would be possible to employ
more standard regression methods for this problem, which would deliver an
estimate of the mean height as a function of the covariates, quantile regression
provides something more: it gives us a sense of what to expect in the spread of
the response variable (height) as a function of the covariates.       

The development of tools for conditional quantile estimation has a rich  
history in both econometrics and statistics. These tools are widely-used for
quantifying uncertainty, and also for characterizing heterogeneous outcomes
across diverse populations. As this is an important problem, many methods to
arrive at quantile estimates abound. It stands to reason that a combination of
different techniques can improve on the accuracy offered by individual base
estimates. Precisely in this vein, the current paper considers the problem of
\emph{model aggregation}, i.e., the task of combining any number of quantile
regression models into a unified estimator. 

To fix notation, let $Y \in \R$ be a response variable of interest, and $X \in
\Xcal$ be an input feature vector used to predict $Y$. A generic way to approach
uncertainty quantification is to estimate the conditional distribution of
$Y|X=x$. However, this can be a formidable challenge, especially in high
dimensions ($\Xcal = \R^d$, where $d$ is large). A simpler alternative is to
estimate conditional quantiles of $Y|X=x$ across a discrete set of quantile
levels $\Tcal \subseteq [0,1]$, that is, to estimate \smash{$g^*(x; \tau) =
F^{-1}_{Y|X=x}(\tau)$} for $\tau \in \Tcal$. Here, for a random variable $Z$
with a cumulative distribution function (CDF) $F_Z$, we denote its level $\tau$
quantile by
\[
F^{-1}_Z(\tau) = \inf\{z : F_Z(z) \geq \tau\}. 
\]
For example, we might choose $\Tcal = \{0.01, 0.02, \ldots, 0.99\}$ to finely 
characterize the spread of $Y | X = x$. 

The aggregation problem can be motivated as follows. Suppose we have a number of
conditional quantile estimates, for instance, through a set of different
quantile regression methods, or various teams submitting their estimates
to a consensus board or as entries in a prediction competition. In all of these
cases, we need an automated strategy to determine which estimate(s) of which  
quantile level(s) should be combined into a consensus model. 

More formally, suppose we have a collection \smash{$\{\hat{g}_j\}_{j=1}^p$} of
conditional quantile models, parametrized by a set of common quantile levels  
$\Tcal$.  Each model \smash{$\hat{g}_j$}, which we refer to as a \emph{base 
model}, provides an estimate of the true conditional quantile function $g^*$. It
is convenient to view $g^*$, and each \smash{$\hat{g}_j$}, as functions from 
$\Xcal$ to $\R^m$, so that $g^*(x)$ outputs a vector of dimension $m = |\Tcal|$,
the number of quantile levels.  We denote the components of this vector by
$g^*(x; \tau)$ for $\tau \in \Tcal$, and similarly for each 
\smash{$\hat{g}_j(x)$}. Given $p \times m$ estimates of $m$ quantile levels by
$p$ base models, each one a function of $x \in \Xcal$, we will study ensemble
estimates, of the generic form:
\[
\hat{g} = H(\hat{g}_1, \ldots, \hat{g}_p) \, : \, \Xcal \to \R^m. 
\] 
In this paper, we focus on linear aggregation procedures $H$, though we allow
the aggregation weights in these linear combinations to be themselves functions
of input feature values $x$, as in: 
\begin{equation}
\label{eq:linear_ensemble}
\hat{g}_w(x)  = \sum_{j=1}^p w_j(x) \cdot \hat{g}_j(x), \quad x \in \Xcal. 
\end{equation}
This form may seem overly restrictive. That said, each term \smash{$\hat{g}_j$}
on its own can be quite powerful. Moreover, as we show later, a sufficiently
flexible parametrization for the weight functions can provide all the modeling
power that we need. In particular, we will consider various aggregation
strategies in which each weight $w_j(x)$ is a scalar, vector, or matrix. The
product ``$\cdot$'' between $w_j(x)$ and \smash{$\hat{g}_j(x)$} in
\eqref{eq:linear_ensemble} is to be interpreted accordingly (more below).  

Our main purpose in what follows is to provide a guided tour of how one might go
about fitting quantile aggregation models of the form
\eqref{eq:linear_ensemble}, of varying degrees of flexibility, and to walk  
through some of the major practical considerations that accompany fitting and
evaluating such models. The aggregation strategies that we consider can be laid
out over the following two dimensions.

\begin{enumerate}
\item \emph{Coarse} versus \emph{medium} versus \emph{fine}: this dimension
  determines the resolution for the parametrization of the weights in 
  \eqref{eq:linear_ensemble}.
  \begin{itemize}
  \item A coarse aggregator uses one weight $w_j$ per base model
    \smash{$\hat{g}_j$}, and we accordingly interpret ``$\cdot$'' in
    \eqref{eq:linear_ensemble} as a scalar-vector product. 
  \item A medium aggregator uses one weight \smash{$\wtj$} per base model  
    \smash{$\hat{g}_j$} and quantile level $\tau$, and we interpret ``$\cdot$''
    in \eqref{eq:linear_ensemble} as a Hadamard (elementwise) product
    between vectors. This allows us to place a higher amount of weight for a
    given model in the tails versus the center the distribution. 
  \item A fine aggregator uses one weight \smash{$\wtnj$} per base model
    \smash{$\hat{g}_j$}, output quantile level $\tau$ (for the output quantile),
    and input quantile level $\nu$ (from a base model), and we interpret
    ``$\cdot$'' in \eqref{eq:linear_ensemble} as a matrix-vector product. This
    allows us to use all of the quantiles from all base models in order to form
    an estimate at a single quantile level for the aggregate model (e.g., the
    aggregate median is informed by all quantiles from all base models, not just
    their medians).    
  \end{itemize}

\item \emph{Global} versus \emph{local}: this dimension determines whether
  or not the weights in \eqref{eq:linear_ensemble} depend on $x$. A global 
  aggregator uses constant weights, $w_j(x) = w_j$ for all $x \in \Xcal$,
  whereas a local aggregator allows these to vary locally (and typically
  smoothly) in $x$. 
\end{enumerate}

Apart from model frameworks, the considerations we give the most attention to
revolve around ensuring \emph{quantile noncrossing}: \smash{$\hat{g}(x; \tau)
\leq \hat{g}(x; \tau')$} for any $x$ and $\tau < \tau'$; and \emph{calibration}:
\smash{$\hat{g}(x; \tau') - \hat{g}(x; \tau)$} contains the response variable
with probability $\approx \tau'-\tau$, at least in some average sense over $x
\in \Xcal$. Finally, we approach all of this work through the lens of the deep
learning, designing methodology to be compatible with standard deep learning
optimization toolkits so as to leverage their convenience of implementation and
scalability. 

\begin{figure}[t]
\includegraphics[width=\textwidth]{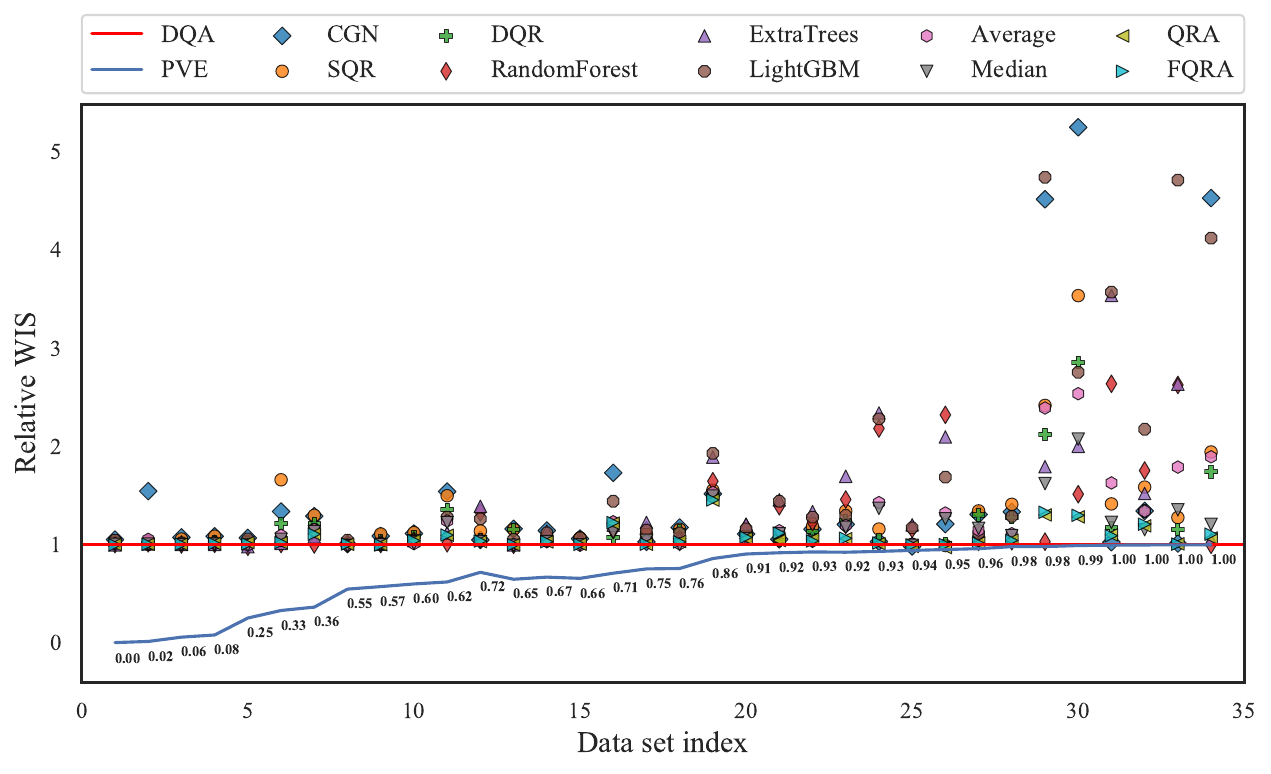}
\captionsetup{format=hang}
\caption{\it\small Weighted interval score (WIS), averaged over out-of-sample
  predictions, for deep quantile aggregation (DQA) and various quantile
  regression methods. DQA is the most flexible aggregation model of the form 
  \eqref{eq:linear_ensemble} that we consider in this paper. The comparison is 
  made over 34 data sets, ordered along the x-axis by proportion of variance 
  explained (PVE), as in \eqref{eq:pve}. The y-axis displays the WIS of each 
  method relative to DQA, so that 1 indicates equal performance to DQA, and a 
  number greater than 1 indicates worse performance than DQA. We can see that
  DQA performs very well overall, especially for larger PVE values (problems
  with higher signal-to-noise ratios).}
\label{fig:intro}
\end{figure}

Before delving into any further details, we present some of our main empirical
results in Figure \ref{fig:intro}. Here and henceforth we use the term
\emph{deep quantile aggregation} (DQA) to refer to the aggregation model in our 
framework with the greatest degree of flexibility, a local fine aggregator where
the local weights are fit using a deep neural network, and we use an adaptive
noncrossing penalty, along with gradient propogation of the min-max sweep  
isotonization operator, to ensure quantile noncrossing (Section 
\ref{sec:noncrossing} provides more details). The figure compares DQA and  
various other quantile regression methods across 34 data sets (Section
\ref{sec:empirical} gives more details). The error metric we use is
\emph{weighted interval score} (WIS), averaged over out-of-sample predictions
(lower WIS is better; more details are given in the next section). 

Shown on the y-axis is the relative WIS of each quantile regression method to
DQA, where 1 indicates equal WIS performance to DQA, and a number greater than 1 
indicates worse performance than DQA. Each point on the x-axis represents a data
set, and we order them by increasing \emph{proportion of variance explained}
(PVE), measured with respect to DQA \smash{$\hat{g}$} (as a proxy for $g^*$) and
test samples $(X^*_i, Y^*_i)$, $i=1,\ldots,m$ by:  
\begin{equation}
\label{eq:pve}
\mathrm{PVE}(\hat{g}) = 1 - \frac{\sum_{i=1}^m (Y^*_i - \hat{g}(X^*_i;  
  0.5))^2}{\sum_{i=1}^m (Y^*_i - \bar{Y}^*)^2}, 
\end{equation}
where \smash{$\hat{g}(x; 0.5)$} denotes the estimate of the conditional median
of $Y|X=x$ from DQA, and \smash{$\bar{Y}^* = \frac{1}{m} \sum_{i=1}^m Y^*_i$} is  
the sample mean of the test responses. The PVE curve itself is drawn in blue on
the figure. The bottom four methods, according to the legend ordering, represent
different aggregators, and the rest are individual base models (some are highly
nonlinear and flexible themselves) that are fed into each aggregation
procedure. As we can see, DQA performs very well across the board, and
particularly for higher PVE values (which we can think of as problems that have
higher signal-to-noise ratios, presenting a greater promise for the flexibility
embodied by DQA), it can provide huge improvements on the base models and other 
aggregators. We should be clear that DQA is not uniformly better than all
methods on all data sets---some points in the figure lie below 1. We provide
an alternate visualization in in Appendix \ref{app:intro_revisit} in which this
is more clearly visible.     

\subsection{Related work}
In this section, we briefly discuss previous work in quantile regression and model aggregation.

\subsubsection{Quantile regression}

Statistical modeling of quantiles dates back to Galton in the 1890s, however,
many facts about quantiles were known long before
\citep{hald1998mathematical}. The modern view on conditional quantile models was
pioneered by Koenker's work on \emph{quantile regression} in the 1970s
\citep{koenker1978regression}; e.g., see \citet{koenker2001quantile,
koenker2005quantile} for nice overviews. This has remained a topic of great
interest, with developments in areas such as kernel machines
\citep{takeuchi2006nonparametric}, additive models \citep{koenker2011additive},
high-dimensional regression \citep{belloni2011penalized}, and graphical models
\citep{ali2016multiple}, just to name a few. Important developments in
distribution-free calibration using quantile regression were given in
\citet{romano2019conformalized, kivaranovic2020adaptive} (which we will return
to later). The rise of deep learning has spurred on new progress in quantile 
regression with neural networks, e.g., \citet{hatalis2017smooth,
dabney2018distributional, xie2019composite, tagasovska2019single,
benidis2020neural}.

\subsubsection{Model aggregation}

Ensemble methods occupy a central place in machine learning (both in theory and
in practice). Seminal work on this topic arose in the 1990s on Bayesian model
averaging, bagging, boosting, and stacking; e.g., see
\citet{dietterich2000ensemble} for a review. While the machine learning
literature has mostly focused on ensembling point predictions, distributional
ensembles have a long tradition in statistics, with a classic reference being
\citet{stone1961opinion}. Combining distributional estimates is also of
great interest in the forecasting community, see \citet{raftery2005using, 
  timmermann2006forecast, ranjan2010combining, gneiting2013combining, 
  gneiting2014probabilistic, kapetanios2015generalised, rasp2018neural,
  cumings2020combination} and references therein.

To the best of our knowledge, the majority of work here has focused on combining
probability densities, and there has been less systematic practical exploration
of how to best combine quantile functions, especially from a flexible
(nonparametric) perspective. \citet{nowotarski2015computing} proposed an
aggregation method they call \emph{quantile regression averaging} (QRA), which
simply performs quantile linear regression on the output of individual
quantile-parametrized base models. Variants of QRA have since been developed, 
for example, factor QRA (FQRA) \citep{maciejowska2016probabilistic} which
applies PCA to reduce dimensionality in the space of base model outputs before
fitting the QRA aggregator. As perhaps evidence for the dearth of sophisticated 
aggregation models\footnote{To be fair, this could also be a reflection of the
  intrinsic difficulty of the forecasting problems in these competitions; for a
  hard problem (low PVE), simple aggregators can achieve competitive performance
  with more complex ones, as seen in Figure \ref{fig:intro}.},        
simple quantile-averaging-type approaches have won various distributional
forecasting competitions \citep{gaillard2016additive, slawek2019forecast,
browell2020quantile}.
Lastly, we note that the study of quantile averaging actually dates back work by
to \citet{vincent1912function}, and hence some literature refers to this method
as \emph{Vincentization}. See also \citet{ratcliff1979group} for relevant
historical discussion. 

\subsection{Outline}

An outline for the remainder of this paper is as follows.

\begin{itemize}
\item Section \ref{sec:background} presents background material on quantile  
  regression, error metrics, and quantile aggregation. This is primarily a
  review of relevant facts from the literature, but we do contribute a few small
  new results, in Propositions \ref{prop:post_hoc_wis} and
  \ref{prop:tail_behavior}.  

\item Section \ref{sec:methods} gives the framework for aggregation methods that 
  we consider in this paper (parametrized by coarse/medium/fine weights on one
  axis, and global/local weights on the other, as explained earlier in the 
  introduction). 

\item Section \ref{sec:noncrossing} investigates methods for ensuring quantile 
  noncrossing while fitting aggregation models, both through explicit
  penalization, and use of differentiable isotonization operators.  

\item Section \ref{sec:conformal} discusses the use of conformal prediction
  (specifically, conformal quantile regression and CV+) to improve calibration 
  post-aggregation. 

\item Section \ref{sec:empirical} provides a broad empirical evaluation of the
  proposed aggregation methods alongside various other aggregators and base
  models. Code to reproduce our all of experimental results is available at: 
  \url{https://github.com/amazon-research/quantile-aggregation}.    
\end{itemize}

\section{Background}
\label{sec:background}
We cover important background material that will help to understand our contributions presented later.

\subsection{Quantile regression and scoring rules} 

We review quantile regression and related scoring rules (error metrics) from the
forecasting literature. 

\subsubsection{Quantile regression}

We begin by recalling the definition of the \emph{pinball loss}, also called the
tilted-$\ell_1$ loss, at a given quantile level $\tau \in [0,1]$. To measure the
error of a level $\tau$ quantile estimate $q$ against an observation $Z$, the
pinball loss is defined by \citep{koenker1978regression}:  
\begin{equation}
\label{eq:pinball}
\pinball_\tau(Z - q) = \begin{cases}
\tau |Z - q| & Z - q \geq 0 \\
(1-\tau) |Z - q| & Z - q < 0.
\end{cases}
\end{equation}
For a continuously-distributed random variable $Z$, the expected pinball loss
$\E[\pinball_\tau(Z - q)]$ is minimized over $q$ at the population-level
quantile \smash{$q^*_\tau =  F_Z^{-1}(\tau)$}. This motivates estimation of 
\smash{$q^*_\tau$} given samples $Z_i$, $i=1,\ldots,n$ by minimizing the sample 
average of the pinball loss: 
\[
\minimize_q \; \frac{1}{n} \sum_{i=1}^n \pinball_\tau(Z_i - q). 
\]
Quantile regression does just this, but applied to the conditional distribution
of $Y|X$. Given samples $(X_i, Y_i)$, $i=1,\ldots,n$, it estimates the true
quantile function \smash{$g^*(x; \tau) = F^{-1}_{Y|X=x}(\tau)$} by solving:
\[
\minimize_{g \in \Gcal} \; \frac{1}{n} \sum_{i=1}^n \pinball_\tau(Y_i - g(X_i)),
\]
over some class of functions $\Gcal$ (possibly with additional regularization in
the above criterion). For example, quantile linear regression takes $\Gcal$ to
be the class of all linear functions of the form $g(x) = x^\T
\beta$. 

As we can see, quantile linear regression is quite a natural extension of
ordinary linear regression---and quantile regression a natural extension of
nonparametric regression more generally---where the focus moves from the
conditional mean to the conditional quantile, but otherwise remains the
same. The pinball loss \eqref{eq:pinball} is convex in $q$ but not
differentiable at zero (unlike the squared loss, associated with mean
estimation), which makes optimization slightly harder. 

To model multiple quantile levels simultaneously, we can simply use
\emph{multiple} quantile regression, where we solve  
\begin{equation}
\label{eq:multi_qr}
\minimize_{g \in \Gcal} \; \frac{1}{n} \sum_{i=1}^n \sum_{\tau \in \Tcal}  
\pinball_\tau(Y_i - g(X_i; \tau)) 
\end{equation}
over a discrete set $\Tcal \subseteq [0,1]$ of quantile levels. We will
generally drop the qualifier ``multiple'', and refer to \eqref{eq:multi_qr} as
quantile regression.  

\subsubsection{Scoring rules}

Another appeal of the pinball loss and the quantile regression framework is its
connection to proper scoring rules from the forecasting literature, which we
detail in what follows.

With $Y$ as our response variable of interest, let $P$ be a predicted
distribution (forecast distribution) and $S$ be a score function, which applied
to $P$ and $Y$, produces $S(P,Y)$. Adopting the notation of
\citet{gneiting2007strictly}, we write $S(P,Q) = \E_Q S(P,y)$, where $\E_Q$
denotes the expectation under $Y \sim Q$. Assuming that $S$ is negatively
oriented (lower values are better), recall that $S$ is said to be \emph{proper}
if $S(P,Q) \geq S(Q,Q)$ for all $P$ and $Q$. As Gneiting and Raftery put it:
``In prediction problems, proper scoring rules encourage the forecaster to make
careful assessments and to be honest.''

For $Y \in \R$, denote by $F$ the cumulative distribution function (CDF)
associated with $P$. The \emph{continuous ranked probability score} (CRPS) is
defined by \citep{matheson1976scoring}:  
\begin{equation}
\label{eq:crps}
\mathrm{CRPS}(F, Y) = \int_{-\infty}^\infty(F(y) - 1\{Y \leq y\})^2 \, \d{y}. 
\end{equation}
This is a well-known proper score, and is popular in various forecasting
communities (e.g., in the atmospheric sciences), in part because it is
considered more robust than the traditional log score.  We also remark that CRPS 
is equivalent to the Cram{\'e}r-von Mises divergence between $F$ and the
empirical CDF $1\{Y \leq \cdot\}$ based on just a single observation $Y$. As
such, it is intimately connected to kernel scores, and more generally, to
integral probability metrics (IPMs) for two-sample testing. For more, see 
Section 5 of \citet{gneiting2007strictly}.

The following reveals an interesting relationship between CRPS \eqref{eq:crps}
and the pinball loss function \eqref{eq:pinball}:    
\begin{equation}
\label{eq:crps_pinball}
\int_{-\infty}^\infty (F(y) - 1\{Y \leq y\})^2 \, \d{y} = 
2 \int_0^1 \pinball_\tau(Y - F^{-1}(\tau)) \, \d\tau. 
\end{equation}
This appears to have been first noted by \citet{laio2007verification}. Their
argument uses integration by parts, but it ignores a few subtleties, so we  
provide a self-contained proof of \eqref{eq:crps_pinball} in Appendix
\ref{app:crps_pinball}.  We can see from \eqref{eq:crps_pinball} that CRPS is
equivalent (up to the constant factor of 2) to an integrated pinball loss,
over all quantile levels $\tau \in [0,1]$. This is quite interesting because
these two error metrics are motivated from very different perspectives, not to
mention different parametrizations (CDF space versus quantile space). 

A natural approximation to the integrated the pinball loss is given by
discretizing, as in:      
\[
\sum_{\tau \in \Tcal} \pinball_\tau(Y - F^{-1}(\tau)),
\]
for a discrete set of quantile levels $\Tcal \subseteq [0,1]$.\footnote{We have
  omitted the adjustment of the summands for spacing between discrete quantile
  levels; note that this only contributes a global scale factor for
  evenly-spaced quantile levels.} 
The interesting connections now continue, in that the above is equivalent to
what is known as the \emph{weighted interval score} (WIS), a proper scoring rule
for forecasts parametrized by a discrete set of quantiles. We assume the
underlying quantile levels are symmetric around 0.5. The collection of
predicted quantiles $F^{-1}(\tau)$, $\tau \in \Tcal$ can then be reparametrized
as a collection of central prediction intervals $(\ell_\alpha, u_\alpha) =
(F^{-1}(\alpha/2), F^{-1}(1-\alpha/2))$, $\alpha \in \Acal$ (each interval here
is parametrized by its exclusion probability), and WIS is defined by:  
\begin{equation}
\label{eq:wis}
\wis(F^{-1}, Y) = \sum_{\alpha \in \Acal} \Big\{\alpha (u_\alpha - \ell_\alpha)
+ 2 \cdot \mathrm{dist}(Y, [\ell_\alpha, u_\alpha]) \Big\}, 
\end{equation}
where $\mathrm{dist}(a, S)$ is the distance between a point $a$ and set $S$ (the 
smallest distance between $a$ and an element of $S$). This scoring metric
appears to have been first proposed in \citet{bracher2021evaluating}, though the 
\emph{interval score} (an individual summand in \eqref{eq:wis}) dates back to
\citet{winkler1972decision}. WIS measures an intuitive combination of sharpness
(first term in each summand) and calibration (second term in each summand). The
equivalence between WIS and pinball loss is now as follows: 
\begin{equation}
\label{eq:wis_pinball}
\sum_{\alpha \in \Acal} \Big\{\alpha (u_\alpha - \ell_\alpha) + 2 \cdot
\mathrm{dist}(Y, [\ell_\alpha, u_\alpha]) \Big\} = 2 \sum_{\tau \in \Tcal} 
\pinball_\tau(Y - F^{-1}(\tau)),
\end{equation}
where \smash{$\Tcal = \cup_{\alpha \in \Acal} \{ \alpha/2, 1-\alpha/2 \}$}. This
is the result of simple algebra and is verified in Appendix
\ref{app:wis_pinball}.   

In summary, we have shown that in training a quantile regression model by
optimizing pinball loss as in \eqref{eq:multi_qr}, we are already equivalently
optimizing for WIS \eqref{eq:wis}, and approximately optimizing for CRPS
\eqref{eq:crps}, where the quality of this approximation improves as the number
of discrete quantile levels increases.

\subsection{Noncrossing constraints and post hoc adjustment} 
\label{sec:noncrossing_background}

An important consideration in fitting conditional quantile models is ensuring
\emph{quantile noncrossing}, that is, ensuring that the fitted estimate
\smash{$\hat{g}$} satisfies:  
\begin{equation}
\label{eq:noncrossing}
\hat{g}(x; \tau) \leq \hat{g}(x; \tau') \quad \text{for all $x$ and $\tau <
  \tau'$}. 
\end{equation}
Two of the most common ways to approach quantile noncrossing are to use
noncrossing constraints during estimation, or to use some kind of post hoc
adjustment rule. In the former approach, we first specify a set $\nocrossx$ at
which we want to enforce noncrossing, and then solve a modified version of
problem \eqref{eq:multi_qr}:
\begin{equation}
\label{eq:multi_qr_constraints}
\begin{alignedat}{2}
&\minimize_{g \in \Gcal} &&\frac{1}{n} \sum_{i=1}^n 
\sum_{\tau \in \Tcal} \pinball_\tau(Y_i - g(X_i; \tau)) \\
&\st \quad && g(x; \tau) \leq g(x; \tau') \quad 
\text{for all $x \in \nocrossx$ and $\tau < \tau'$},
\end{alignedat}
\end{equation}
as considered in \citet{takeuchi2006nonparametric, dette2008non, 
bondell2010noncrossing}, among others (and even earlier in
\citealt{fung2002knowledge} in a different context). The simplest choice is to
take \smash{$\nocrossx = \{X_i\}_{i=1}^n$}, so as to enforce noncrossing at the
training feature values; but in a transductive setting where we have unlabeled
test feature values at training time, these could naturally be included in
$\nocrossx$ as well.

The latter strategy, post hoc adjustment, has been studied in
\citet{chernozhukov2010quantile, kuleshov2018accurate, song2019distribution}
(and even earlier in \citealt{le2006simpler} in a different context). In this 
approach, we solve the original multiple quantile regression problem
\eqref{eq:multi_qr}, but then at test time, at any input feature value $x \in
\Xcal$, we output
\begin{equation}
\label{eq:post_hoc}
\tilde{g}(x) = \Scal(\hat{g}(x)),
\end{equation}
where $\Scal : \R^m \to \iso^m$ is a user-chosen isotonization operator. Here,
recall $m = |\Tcal|$ is the number of discrete quantile levels, and $\iso^m = \{
v \in \R^m : v_i \leq v_{i+1}, \, i=1,\ldots,m-1 \}$ denotes the isotonic cone
in $m$ dimensions. Two widely-used isotonization operators are \emph{sorting}:  
\begin{equation}
\label{eq:sort}
\sort(v) = (v_{(1)}, \ldots, v_{(m)}),
\end{equation}
where we use the classic order statistic notation (here \smash{$v_{(i)}$}
denotes the $i\th$ largest element of $v$), and \emph{isotonic projection}: 
\begin{equation}
\label{eq:iso_proj}
\isoproj(v) = \argmin_{u \in \iso^m} \; \|v - u\|_2.
\end{equation}
The set $\iso^m$ is a closed convex cone, which means the $\ell_2$ projection
operator onto $\iso^m$ is well-defined (the minimization in \eqref{eq:iso_proj}
has a unique solution), and well-behaved (it is nonexpansive, i.e., Lipschitz
continuous with Lipschitz constant $L=1$, and therefore almost everywhere 
differentiable). Furthermore, there are fast linear-time algorithms for isotonic
projection, such as the famous \emph{pooled adjacent violators algorithm}
(PAVA) of \citet{barlow1972statistical}. Note that while $\sort(v) \in \iso^m$,
this does not mean that $\|\sort(v) - v\|_2$ is the shortest distance between
$v$ and $\iso^m$. As such, the solutions to \eqref{eq:sort} and
\eqref{eq:iso_proj} differ in general.   

The constrained approach \eqref{eq:multi_qr_constraints} has the advantage that
the constraints offer a form of regularization and for this reason, may help
improve accuracy over post hoc techniques. It has the disadvantage of increasing 
the computational burden (depending on the nature of the model class $\Gcal$,
these constraints can actually be highly nontrivial to incorporate), and of only
ensuring noncrossing over some prespecified finite set $\nocrossx$. By
comparison, the post hoc approach \eqref{eq:post_hoc} is computationally trivial
(in the case of sorting \eqref{eq:sort} and isotonic projection
\eqref{eq:iso_proj}), and by construction, ensures noncrossing at each $x \in 
\Xcal$.

Moreover, the post hoc approach is simple enough that it is possible to prove 
some general guarantees about its effect. The following is a transcription of
some important results along these lines from the literature, for sorting and
isotonic projection. 

\begin{proposition}[\citealt{chernozhukov2010quantile, robertson1998order}] 
\label{prop:post_hoc_lp}
Let $\Tcal \subseteq [0,1]$ be an arbitrary finite set, and denote by 
\[
g^*(x) = \big\{g^*(x; \tau)\big\}_{\tau \in \Tcal} \quad \text{and} \quad 
\hat{g}(x) = \big\{\hat{g}(x; \tau)\big\}_{\tau \in \Tcal}
\]
an arbitrary true and estimated conditional quantile function at a point $x$
(that is, $g^*(x)$ and \smash{$\hat{g}(x)$} are arbitary vectors in $\iso^m$ and
$\R^m$, respectively, where $m = |\Tcal|$). Then the following holds for the
post-adjusted estimate \smash{$\tilde{g}(x)$} in \eqref{eq:post_hoc}.  

\begin{itemize}
\item[(i)] If $\Scal = \sort$, the sorting operator \eqref{eq:sort}, then the
  $\ell_p$ norm error between the estimate and $g^*(x)$ can only improve for any 
  $p \geq 1$, that is, \smash{$\|\tilde{g}(x) - g^*(x)\|_p \leq  \|\hat{g}(x) -
    g^*(x)\|_p $} for any $p \geq 1$. Moreover, if sorting is nontrivial:
  \smash{$\tilde{g}(x) \not= \hat{g}(x)$}, and $p > 1$, then the $\ell_p$ error  
  inequality is strict.   

\item[(ii)] If $\Scal = \isoproj$, the isotonic projection operator
  \eqref{eq:iso_proj}, then the same result holds as in part (i).
\end{itemize}
\end{proposition}

Part (i) of this proposition is due to \citet{chernozhukov2010quantile} 
and is an application of the classical rearrangement inequality 
\citep{hardy1934inequalities}. Part (ii) is due to
\citet{robertson1998order}. 

In our ensemble setting, metrics on predictive accuracy are more
relevant. Towards this end, the next proposition contributes a new but small 
result on post hoc adjustment and WIS.     

\begin{proposition}
\label{prop:post_hoc_wis}
As in the last proposition, let $\Tcal \subseteq [0,1]$ be an arbitrary finite
set, and \smash{$\hat{g}(x) = \{\hat{g}(x; \tau)\}_{\tau \in \Tcal}$} be an
estimate of the conditional quantile function at a point $x$. Then the following
holds for the post-adjusted estimate \smash{$\tilde{g}(x)$} in
\eqref{eq:post_hoc}, and for any $y \in \R$.

\begin{itemize}
\item[(i)] If $\Scal = \sort$, the sorting operator \eqref{eq:sort}, then
  the pinball loss can only improve:
  \[
  \sum_{\tau \in \Tcal} \pinball_\tau(y - \tilde{g}(x; \tau)) \leq 
  \sum_{\tau \in \Tcal} \pinball_\tau(y - \hat{g}(x; \tau)). 
  \]
  When $\Tcal$ is symmetric around 0.5, this means 
  \smash{$\wis(\tilde{g}(x), y) \leq \wis(\hat{g}(x), y)$} as well (by the 
  equivalence between pinball loss and WIS in \eqref{eq:wis_pinball}). 
  Moreover, if sorting is nontrivial: \smash{$\tilde{g}(x) \not=
    \hat{g}(x)$}, then the pinball or WIS improvement is strict.   

\item[(ii)] If $\Scal = \isoproj$, the isotonic projection operator
  \eqref{eq:iso_proj}, then the same result holds as in part (i).
\end{itemize}
\end{proposition}

The proof of Proposition \ref{prop:post_hoc_wis} elementary and given in
Appendix \ref{app:post_hoc_wis}. Interestingly, the proof makes use of
particular algorithms for sorting and isotonic projection (bubble sort and PAVA,
respectively).

Note that, as the inequalities in Propositions \ref{prop:post_hoc_lp} and
\ref{prop:post_hoc_wis} hold pointwise for each $x \in \Xcal$, they also hold in 
an average (integrated) sense, with respect to an arbitrary distribution on
$\Xcal$. Later in Section \ref{sec:noncrossing}, we compare and combine these
and other noncrossing strategies, through extensive empirical
evaluations. Analogous to noncrossing constraints in
\eqref{eq:multi_qr_constraints}, we consider a crossing penalty (which is
similar, but more computationally efficient); and as for rules like sorting and
isotonic projection, we evaluate them not only post hoc, but also as layers in
training. 

\subsection{Quantile versus probability aggregation}
\label{sec:quantile_vs_probability}

When it comes to combining uncertainty quantification models, we have a number
of options: we can either average over probabilities or over quantiles. These
strategies are quite different, and often lead to markedly different
outcomes. This subsection provides some theoretical background comparing the two
strategies. We see it as important to review this material because it is a
useful guide to thinking about quantile aggregation, which is likely less
familiar to most readers (than probability aggregation). 

In this subsection only, the term \emph{average} refers to a weighted linear
combination where the weights are nonnegative and sum to 1. For each
$j=1,\ldots,p$, let $F_j$ be a cumulative distribution function (CDF);
$f_j=F_j'$ be its probability density function; let \smash{$Q_j = F^{-1}_j$}
denote the corresponding quantile function; and $q_j=Q_j'$ the quantile density
function. A standard fact that relates these objects:  
\begin{equation}
\label{eq:probability_quantile}
q_j(u) = \frac{1}{f_j(Q_j(u))} \quad \text{and} \quad 
f_j(v) = \frac{1}{q_j(F_j(v))}.  
\end{equation}
The first fact can be checked by differentiating $Q_j(F_j(v)) = v$, applying the
chain rule, and reparametrizing via $u=F_j(v)$. The second follows similarly via
$F_j(Q_j(u)) = u$.

We compare and contrast two ways of averaging distributions. The first way is in
probability space, where we define for weights $w_j \geq 0$, $j=1,\ldots,p$ such
that \smash{$\sum_{j=1}^p w_j = 1$}, 
\[
F = \sum_{j=1}^p w_j F_j.
\]
The associated density is simply \smash{$f = \sum_{j=1}^p w_j f_j$} since
differentiation is a linear operator. The second way to average is in quantile
space, defining 
\[ 
\bar{Q} = \sum_{j=1}^p w_j Q_j,
\]
where now $\bar{q} = \sum_{j=1}^p w_j q_j$ is the associated quantile density, 
again by linearity of differentiation. Denote the CDF and probability density
associated with the quantile average by \smash{$\bar{F} = \bar{Q}^{-1}$}, and
\smash{$\bar{f} = \bar{F}'$}. Note that from \eqref{eq:probability_quantile},
we can reason that \smash{$\bar{f}$} is a highly \emph{nonlinear} function of
$f_j$, $j=1,\ldots,p$.     

\begin{figure}[htb]
\centering
\includegraphics[width=0.9\textwidth]{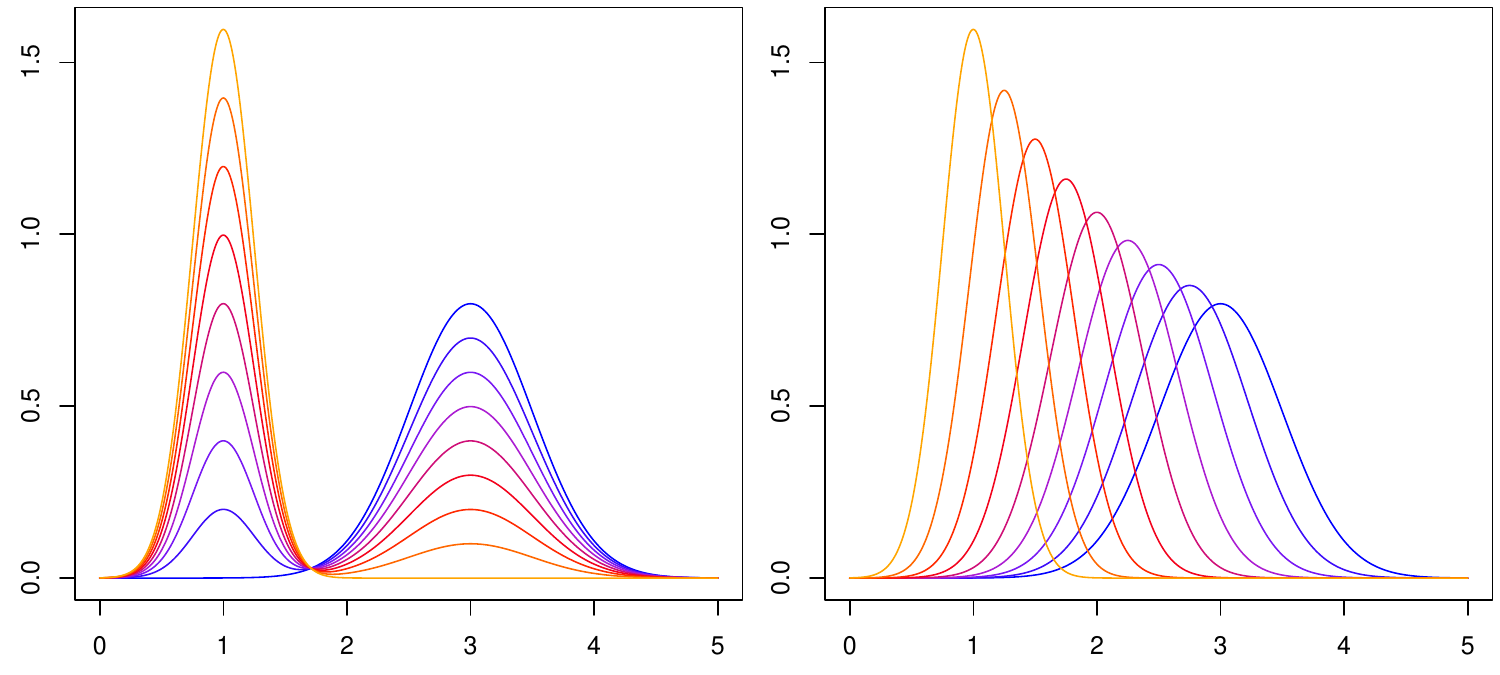}
\captionsetup{format=hang}
\caption{\it\small Densities that result from a probability average (left) or
quantile average (right) of two normal distributions $N(1,0.25^2)$ and
$N(3, 0.5^2)$, as the weight on the first density varies from 1 (orange) to 0
(blue).} 
\label{fig:probability_vs_quantile}
\end{figure}

A simple example can go a long way to illustrate the differences between the
distributions resulting from probability and quantile averaging. In Figure
\ref{fig:probability_vs_quantile}, we compare these two ways of averaging on a
pair of normal distributions with different means and variances.  Here we see
that probability averaging produces the familiar mixture of normals, which is
bimodal. The result of quantile averaging is very different: it is always
unimodal, and instead of interpolating between the tail behaviors of $f_1$ and  
$f_2$ (as $f$ does), it appears that \emph{both} tails of \smash{$\bar{f}$} are
generally thinner than those of $f_1$ and $f_2$.

It seems that quantile averaging is doing something that is both like
translation and scaling in probability density space. Next we explain this
phenomenon precisely by recalling a classic result.   

\subsubsection{Shape preservation}

An aggregation procedure $H$ is said to be \emph{shape-preserving} if, for any
location-scale family $\Pcal$ (such as Normal, t, Laplace, or Cauchy) whose
elements differ only by scale and location parameters, we have    
\[ 
F_j \in \Pcal, \, j=1,\ldots,p \implies F = H(F_1,\ldots,F_p) \in \Pcal.  
\]
Probability averaging is clearly not shape-preserving, however, interestingly,
quantile averaging is: if each $F_j$ a member of the same location-scale family
with a base CDF $L$, then we can write $F_j(v) = L((v-\theta_j)/\sigma_j)$, thus
$Q_j(u) = \theta_j + \sigma_j L^{-1}(u)$, so \smash{$\bar{Q}$} is still of the
form $\theta + \sigma L^{-1}$ and \smash{$\bar{F}$} is also in the
location-scale family. The next proposition collects this and related results
from the literature.

\begin{proposition}[\citealt{thomas1980appropriate, genest1992vincentization}]
\label{prop:shape_preservation} \hfill 
\begin{itemize}
\item[(i)] Quantile averaging is shape-preserving.        
\item[(ii)] Location-scale families $\Pcal$ are the only ones with respect to
  which quantile averaging is a closed operation (meaning $F_j \in \Pcal$,
  $j=1,\ldots,p$ implies \smash{$\bar{F} \in \Pcal$}).    
\item[(iii)] Quantile averaging is the only aggregation procedure $H$, of those
  satisfying (for $h$ not depending on $y$): 
  \[
    H(F_1,\ldots,F_p)^{-1}(u) = h(Q_1(u),\ldots,Q_p(u)),
  \]
  that is shape-preserving.
\end{itemize}
\end{proposition}

Part (ii) is due to \citet{thomas1980appropriate}. Part (iii) is due to
\citet{genest1992vincentization}, following from an elegant application of
Pexider's equation.

The parts of Proposition \ref{prop:shape_preservation}, taken together, suggest
that quantile averaging is somehow ``tailor-made'' for shape preservation in a
location-scale family---which can be seen as either a pro or a con, depending on
the application one has in mind. To elaborate, suppose that in a quantile
regression ensembling application, each base model outputs a normal distribution
for its predicted distribution at each $x$ (with different means and
variances). If the normal assumption is warranted (i.e., it actually describes
the data generating distribution) then we would want our ensemble to retain 
normality, and quantile averaging would do exactly this. But if the normal
assumption is used only as a working model, and we are looking to combine base
predictions as a way to construct some flexible and robust model, then the
shape-preserving property of quantile averaging would be problematic. In
general, to model arbitrary distributions without imposing strong assumptions,
we are therefore driven to use linear combinations of quantiles that allow 
\emph{different aggregation weights to be used for different quantile levels}, 
of the form \smash{$\bar{Q}(u) = \sum_{j=1}^p w_j(u) Q_j(u)$}. 

\subsubsection{Moments and sharpness}

We recall an important result about moments of the distributions returned by
probability and quantile averages. For a distribution $G$, we denote its
uncentered moment of order $k \geq 1$ by \smash{$m_k(G) = \E_G[X^k]$}, where the
expectation is under $X \sim G$.

\begin{proposition}[\citealt{lichtendahl2013better}]
\label{prop:moments_sharpness} \hfill
\begin{itemize}
\item[(i)] A probability and quantile average always have equal means:
  \smash{$m_1(F) = m_1(\bar{F})$}.  
\item[(ii)] A quantile average is always sharper than a probability average:
  \smash{$m_k(\bar{F}) \leq m_k(F)$} for any even $k \geq 2$.   
\end{itemize}
\end{proposition}

Note that sharpness is only a desirable property if it does not come at the
expense of calibration. With this in mind, the above result cannot be understood
as a pro or con of quantile averaging without any context on calibration. That
said, the relative sharpness of quantile averages to probability averages is an
important general phenomenon to be aware of. 

\subsubsection{Tail behavior}

Lastly, we study the action of quantile averaging on the tails of the subsequent
probability density. Simply starting from \smash{$\bar{q} = \sum_{j=1}^p w_j
  q_j$}, differentiating, and using \eqref{eq:probability_quantile}, we get
\[
\frac{1}{\bar{f}(\bar{Q}(u))} = \sum_{j=1}^p \frac{w_j}{f_j(Q_j(u))}. 
\]
That is, the probability density \smash{$\bar{f}$} at the level $u$ quantile is
a (weighted) \emph{harmonic mean} of the densities $f_j$ at their respective
level $u$ quantiles. Since harmonic means are generally (much) smaller than
arithmetic means, we would thus expect \smash{$\bar{f}$} to have thinner tails
than $f$. The next result formalizes this. We use $g(v) = o(h(v))$ to mean
$g(v)/h(v) \to 0$ as $v \to \infty$, and $g(v) \asymp h(v)$ to mean $g(v)/h(v)
\to c \in (0, \infty)$ as $v \to \infty$.

\begin{proposition}
\label{prop:tail_behavior}
Assume that $p=2$, $f_2(v) = o(f_1(v))$, and the weights $w_1,w_2$ are
nontrivial (they lie strictly between 0 and 1). Then the density from
probability averaging satisfies $f(v) \asymp f_1(v)$. Assuming further that
$f_1$ is log-concave, the density from quantile averaging satisfies
\smash{$\bar{f}(v) = o(f_1(v))$}.  
\end{proposition}

The proof is based on \eqref{eq:probability_quantile}, and is deferred to
Appendix \ref{app:tail_behavior}. The assumption that $f_1$ is log-concave for
the quantile averaging result is stronger than it needs to be (as is the
restriction to $p=2$), but is used to simplify exposition. Proposition 
\ref{prop:tail_behavior} reiterates the importance of allowing for
level-dependent weights in a linear combination of quantiles. For applications
in which there is considerable uncertainty about extreme events (especially ones
in which there is disagreement in the degree of uncertainty between individual
base models), we would not want an ensemble to de facto inherit a particular
tail behavior---whether thin or thick---but want to endow the aggregation
procedure with the ability to adapt its tail behavior as needed.  

\section{Aggregation methods}
\label{sec:methods}

In what follows, we describe various aggregation strategies for combining
multiple conditional quantile base models into a single model.  Properly
trained, the ensemble should be able to account for the strengths and weaknesses
of the base models and hence achieve superior accuracy to any one of them. This
of course will only be possible if there is enough data to statistically
identify such weaknesses and enough flexibility in the aggregation model to
adjust for them.

\subsection{Out-of-sample base predictions}
\label{sec:oos_base}

To avoid overfitting, a standard model ensembling scheme uses \emph{out-of-fold
  predictions}  from all base models when learning ensemble weights; see, e.g., 
\citet{van2007super, erickson2020auto}. As in cross-validation, here we 
randomly partition the training data \smash{$\{(X_i,Y_i)\}_{i=1}^n$} into $K$
disjoint and equally-sized folds (all of our experiments use $K=5$), where the 
folds \smash{$\{\Ical_k\}_{k=1}^K$} form a partition of the index set
$\{1,\ldots,n\}$. For each fold $\Ical_k$, we retrain each base model on the
other folds $\{\Ical_1,\ldots,\Ical_K\} \setminus \{\Ical_k\}$ to obtain an
out-of-sample prediction at each $X_i$, $i \in \Ical_k$, denoted
\smash{$\hat{g}_j^{-k(i)}(X_i)$} (where we use $k(i)$ for the index of the  
fold containing the $i\th$ data point, and the superscript $-k(i)$ on the base
model estimate indicates that the model is trained on folds excluding the $i\th$ 
data point).  When fitting the ensemble weights, we only consider quantile
predictions from each model that are out-of-sample:
\begin{equation}
\label{eq:linear_ensemble_oos}
\hat{g}_w(X_i)  = \sum_{j=1}^p w_j(X_i) \cdot \hat{g}_j^{-k(i)}(X_i), \quad  
i=1,\ldots,n. 
\end{equation}
Once the ensemble weights have been learned, we make predictions at any new test
point $x \in \Xcal$ via \eqref{eq:linear_ensemble}, where the base models 
\smash{$\{\hat{g}_j\}_{j=1}^p$} have been fit to the full training set
\smash{$\{(X_i,Y_i)\}_{i=1}^n$}.

\subsection{Global aggregation weighting schemes}
\label{sec:global_aggr}

For now, we assume that each weight is constant with respect to $x \in \Xcal$,
i.e., $w_j(x) = w_j$ for all $j=1,\ldots,p$. This puts us in the category of
\emph{global} aggregation procedures; we will turn to \emph{local} aggregation
procedures in the next subsection. All strategies described below can be cast in
the following general form. We fit the global aggregation weights $w$ by solving
the optimization problem:
\begin{equation}
\label{eq:linear_ensemble_opt}
\begin{alignedat}{2}
&\minimize_w &&\frac{1}{n} \sum_{i=1}^n 
\sum_{\tau \in \Tcal} \pinball_\tau\bigg( Y_i - \sum_{j=1}^p w_j \cdot
\hat{g}_j^{-k(i)}(X_i; \tau) \bigg) \\   
&\st \quad && Aw = 1, \quad w \geq 0, 
\end{alignedat}
\end{equation} 
Note that each base model prediction used in \eqref{eq:linear_ensemble_opt} is
an out-of-fold prediction, as in \eqref{eq:linear_ensemble_oos}. Moreover, $A$
is a linear operator that encodes the unit-sum constraint on the weights
(further details on the parametrization for each case is described below):  
\[
Aw = 
\begin{cases}
  \sum_{j=1}^p w_j \vphantom{\big\{\sum_{j=1}^p} & \text{coarse case} \\   
  \big\{\sum_{j=1}^p \wtj\big\}_{\tau \in \Tcal} & \text{medium case} \\ 
  \big\{\sum_{j=1}^p \sum_{\nu \in \Tcal} \wtnj\big\}_{\tau \in \Tcal} &
  \text{fine case}. 
\end{cases}
\]
Lastly, in the second line of \eqref{eq:linear_ensemble_opt}, we use 1 to denote
the vector of all 1s (of appropriate  dimension), and the constraint $w \geq 0$
is to be interpreted elementwise.     

Within this framework, we can consider various weighted ensembling strategies of 
increasing flexibility (akin to coffee grind sizes). These were covered briefly
in the introduction, and we expand on them below.  

\begin{itemize} 
\item \emph{Coarse.} To each base model \smash{$\hat{g}_j$}, we allocate a 
  weight $w_j$, shared over all quantile levels. With $p$ base models, a coarse
  aggregator learns $p$ weights, satisfying \smash{$\sum_{j=1}^p w_j = 1$}. The
  product ``$\cdot$'' in \eqref{eq:linear_ensemble_opt} is just a scalar-vector
  product, so that ensemble prediction for level $\tau$ is   
  \[
  \sum_{j=1}^p \big(w_j \cdot \hat{g}_j(x)\big)_\tau = \sum_{j=1}^p w_j \,
  \hat{g}_j(x; \tau). 
  \]
  This appears to be the standard way to build weighted ensembles, including
  quantile ensembles, see, e.g., \citet{nowotarski2018electricity,
    browell2020quantile, zhang2020load, uniejewski2021regularized}. 
  However, it has clear limitations in the quantile setting, as outlined in 
  Section \ref{sec:quantile_vs_probability}. To recap, coarsely-aggregated 
  distributions will generally have thinner tails than the thickest of the base
  model tails (Proposition \ref{prop:tail_behavior}), and if all base models
  produce quantiles according to some common location-scale family, then the
  coarsely-aggregated distribution will remain in this family (Proposition
  \ref{prop:shape_preservation}). Medium and fine strategies, described next, 
  are free of such restrictions. 

\item \emph{Medium.} To each base model \smash{$\hat{g}_j$} and quantile level 
  $\tau$, we allocate a weight \smash{$\wtj$}. With $p$ base models and $m$
  quantile levels, a medium aggregator learns $p \times m$ weights, satisfying
  \smash{$\sum_{j=1}^p \wtj = 1$} for each quantile level $\tau$. The weight
  assigned to each base model is a vector $w_j \in \R^p$, and the product
  ``$\cdot$'' in \eqref{eq:linear_ensemble_opt} is the Hadamard (elementwise)
  product between vectors, so that ensemble prediction for level $\tau$ is  
  \[
  \sum_{j=1}^p \big(w_j \cdot \hat{g}_j(x)\big)_\tau = \sum_{j=1}^p \wtj \,
  \hat{g}_j(x; \tau). 
  \]
  This approach enables the ensemble to account for the fact that different base  
  models may be better or worse at predicting certain quantile levels. It has
  been considered in quantile aggregation by, e.g.,
  \citet{nowotarski2015computing, maciejowska2016probabilistic, lima2017out,   
    tibshirani2020tools}. 

\item \emph{Fine.} To each base model \smash{$\hat{g}_j$}, output (ensemble)
  quantile level $\tau$, and input (base model) quantile level $\nu$, we
  allocate a weight \smash{$\wtnj$}. Thus with $p$ base models and $m$
  quantile levels, a fine aggregator learns $p \times m \times m$ weights,
  satisfying \smash{$\sum_{j=1}^p \sum_{\nu \in \Tcal} \wtnj = 1$} for each
  quantile level $\tau$. The weight assigned to each base model is a matrix $w_j
  \in \R^{m \times m}$, and ``$\cdot$'' in \eqref{eq:linear_ensemble_opt} is a
  matrix-vector product, so the ensemble prediction for level $\tau$ is     
  \[
  \sum_{j=1}^p \big(w_j \cdot \hat{g}_j(x)\big)_\tau = \sum_{j=1}^p \sum_{\nu
    \in \Tcal} \wtnj \, \hat{g}_j(x; \nu).  
  \]
  This strategy presumes that for a given quantile level $\tau$, base model
  estimates for other quantile levels $\nu \not= \tau$ could also be useful for 
  aggregation purposes. Note that this is particularly pertinent to a setting
  which some base models are poorly calibrated or produce unstable estimates for
  one particular quantile level. To the best of our knowledge, this type of
  aggregation is not common and has not been thoroughly studied in the
  literature.    
\end{itemize}

As a way of comparing the flexibility offered by the three strategies, denote
the ensemble prediction at $x$ by \smash{$\hat{g}_w(x) = \sum_{j=1}^p \big(w_j  
  \cdot \hat{g}_j(x)\big)_\tau$}, and note that (where we abbreviate $\range_{s
  \in S} a_s = [\min_{s \in S} a_s, \, \max_{s \in S} b_s]$):
\[
\hat{g}_w(x; \tau) \in \begin{cases} 
\range_{j=1,\ldots,p} \, \hat{g}_j(x; \tau) & \text{coarse and medium cases} \\
\range_{j=1,\ldots,p, \, \nu \in \Tcal} \, \hat{g}_j(x; \nu) & \text{fine case}.
\end{cases} 
\]
In the medium and fine cases, and any element in the respective ranges above
is achieveable (by varying the weights appropriately). In the coarse case, this
is not true, and \smash{$\hat{g}_w$} is restricted by the shape of the
individual quantile functions (recall Section
\ref{sec:quantile_vs_probability}).       

Furthermore, it is not hard to see that problem \eqref{eq:linear_ensemble_opt}
is equivalent to a linear program (LP), and can be solved with any standard
convex optimization toolkit. While conceptually tidy, solving this LP
formulation can be overly computationally intensive at scale. In this paper, 
rather than relying on LP formulations, we fit global aggregation weights by
optimizing \eqref{eq:linear_ensemble_opt} via stochastic gradient descent (SGD)
methods, which are much more scalable to large data sets through their use of
subsampled mini-batches and backpropagation in deep learning toolkits that can
leverage hardware accelerators (GPUs). Outside of computational efficiency, the
use of SGD also provides implicit regularization to avoid overfitting (which we
can further control using early stopping). To circumvent the simplex constraints
in \eqref{eq:linear_ensemble_opt}, we reparametrize the weights
\smash{$\{w_j\}_{j=1}^p$} using a softmax layer applied to (unconstrained) 
parameters \smash{$\{\phi_j\}_{j=1}^p$}, which for the coarse case we can 
write as:
\[
w = \softmax(\phi) = \bigg\{\frac{e^{\phi_j}}{\sum_{\ell=1}^p e^{\phi_\ell}}
\bigg\}_{j=1}^p.  
\]
The other cases (medium and fine) follow similarly. 

Yet another advantage of using SGD and deep learning toolkits to solve
\eqref{eq:linear_ensemble_opt} is that this extends more fluidly to the setting
of local aggregation weights, which we described next. 

\subsection{Local aggregation via neural networks} 
\label{sec:local_aggr}

To fit \emph{local} aggregation weights that vary with $x \in \Xcal$, we use a
neural network approach. For concreteness, we describe the procedure in the
context of fine aggregation weights, and the same idea applies to the other
cases (coarse and medium) as well. We will refer to the local-fine approach,
described below, as \emph{deep quantile aggregation} or DQA. 

To fit weight functions \smash{$\wtnj : \Xcal \to \R$}, we model these jointly
over all base models $j \in \{1,2, \ldots, p\}$ and all pairs of quantile levels
$(\tau, \nu) \in \Tcal \times \Tcal$ using a single neural network $G$. All but
the last layer of $G$ forms a feature extractor $f_\theta: \Xcal \rightarrow
\R^d$ that maps input feature values $x \in \Xcal$ to a hidden representation $h
\in \R^d$. In our experiments, we take $f_\theta$ to be a standard feed-forward
network parametrized by $\theta$. The last layer of $G$ is defined, for each
output quantile level $\tau \in \Tcal$, by a matrix $W^\tau \in \R^{pm \times
  d}$ that maps the hidden representation $h$, followed by a softmax operation
(to ensure the simplex constraints are met), to fine aggregation weights
\smash{$w^\tau(x) \in \R^{pm}$}, as in:    
\begin{equation} 
\label{eq:local_aggr}
w^\tau(x) = G(x; \theta, W^\tau) = \softmax\big( W^\tau f_\theta(x) \big), \quad
\tau \in \Tcal.  
\end{equation}
The parameters $\theta$ and \smash{$\{W^\tau\}_{\tau \in \Tcal}$} can be jointly 
optimized via SGD, applied to the optimization problem:
\begin{equation}
\label{eq:linear_ensemble_opt_local}
\minimize_{\theta, \, \{W^\tau\}_{\tau \in \Tcal}} \; \frac{1}{n} \sum_{i=1}^n  
\sum_{\tau \in \Tcal} \pinball_\tau\bigg( Y_i - \sum_{j=1}^p \sum_{\nu \in
  \Tcal} G(x; \theta, W^\tau)_{j,\nu} \, \hat{g}^{-k(i)}_j(X_i; \nu) \bigg).
\end{equation} 
Note that there are no explicit constraints because the softmax parametrization
in \eqref{eq:local_aggr} implicitly fulfills the appropriate constraints. 

Why would we want to move from using global aggregation weights
\eqref{eq:linear_ensemble_opt}, which do not depend on the feature values $x$,
to using local aggregation weights \eqref{eq:local_aggr},
\eqref{eq:linear_ensemble_opt_local}, which do? Aside from the general
perspective of increasing model flexibility, local weights can be motivated by
the observation that, in any given problem setting, it may well be the case that
that certain constituent models perform better (output more accurate quantile
estimates) in some parts of the feature space, while others perform better in
other parts. In such cases, a local aggregation scheme will be able to adapt 
its weights accordingly, whereas a global scheme will not, and will be stuck
with attributing some global level of preference between the models.

\subsection{Interpreting the local aggregation scheme}

We can interpret our proposal in \eqref{eq:local_aggr},
\eqref{eq:linear_ensemble_opt_local} as a \emph{mixture of experts} ensemble
\citep{jacobs1991adaptive} where the gating network $G$ emits predictions about
which ``experts'' will be most accurate for any particular $x$. In the
local-fine setting (DQA), the ``experts'' correspond to estimates of individual
quantiles from individual base models, and we define a separate gating scheme
for each target quantile level $\tau$.

We can also view our local-fine aggregator through the lens of an
\emph{attention} mechanism \citep{bahdanau2015neural}, which adaptively attends
to the different quantile estimates from each base model, and uses a different
attention map for each $\tau$ and each $x$. In terms of the architecture design,
we use a representation of the tuple $(\tau, x)$ both as a \emph{key} and as a
\emph{query}. The individual quantiles \smash{$\hat{g}_j(x; \tau)$} are then
used as \emph{values} in the (query, key, value) mechanism commonly used.    

\begin{figure}[p]
\includegraphics[width=\textwidth]{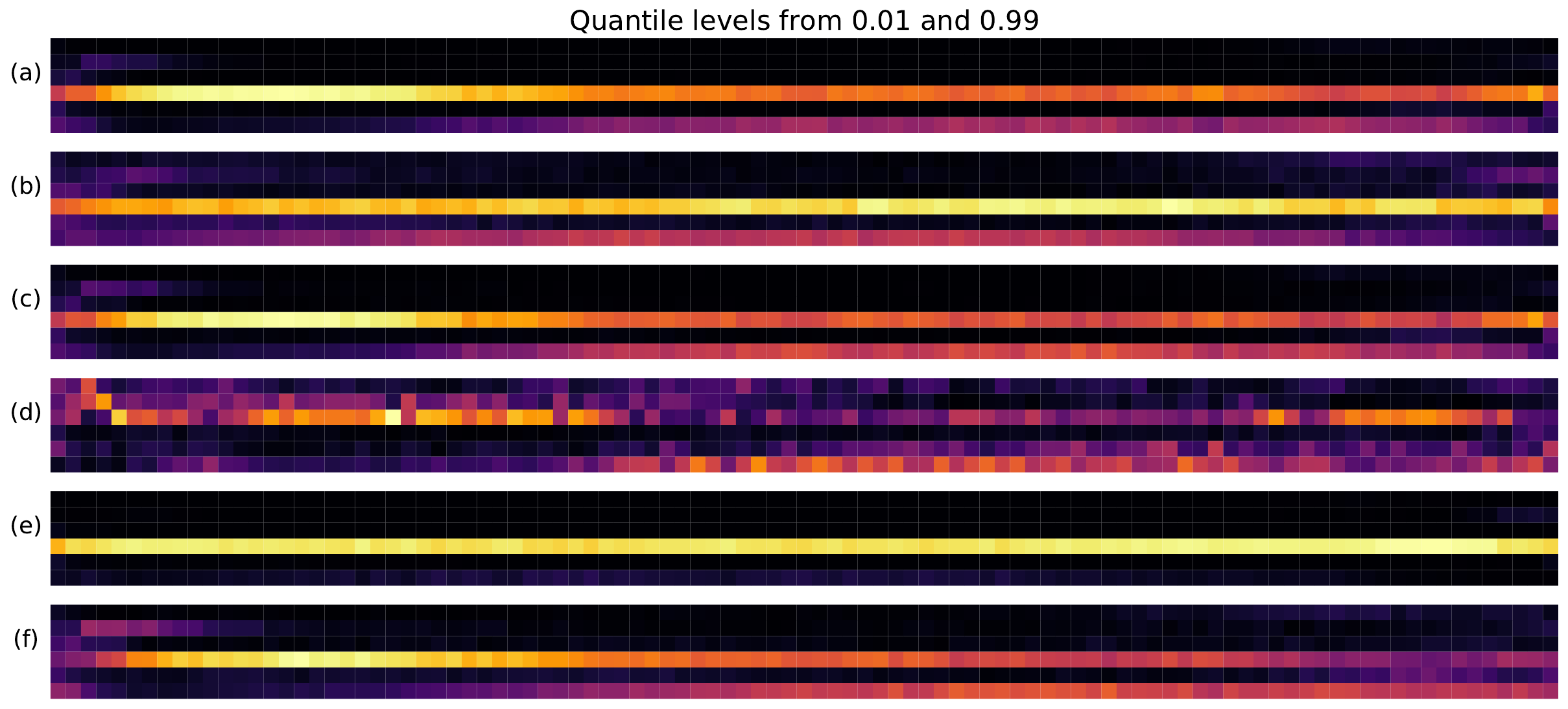}
\captionsetup{format=hang}
\caption{\it\small Heatmaps of aggregation weights fit by the local-medium 
  aggregator to the concrete data set, at 6 different input feature values. Each
  heatmap in (a)--(f) corresponds to a particular input feature value $x$. Its
  rows correspond to the $p=6$ base models, and columns to the $m=99$ 
  quantile levels. A black color means that a given base prediction is
  essentially ignored by the aggregator. We can see that for different
  $x_i$ the algorithm chooses different base models but also that it
  then mostly uses the estimates of these base models in a consistent fashion
  across quantile levels.} 
\label{fig:local_medium_heatmap}

\bigskip\bigskip
\centering
\includegraphics[width=0.49\textwidth]{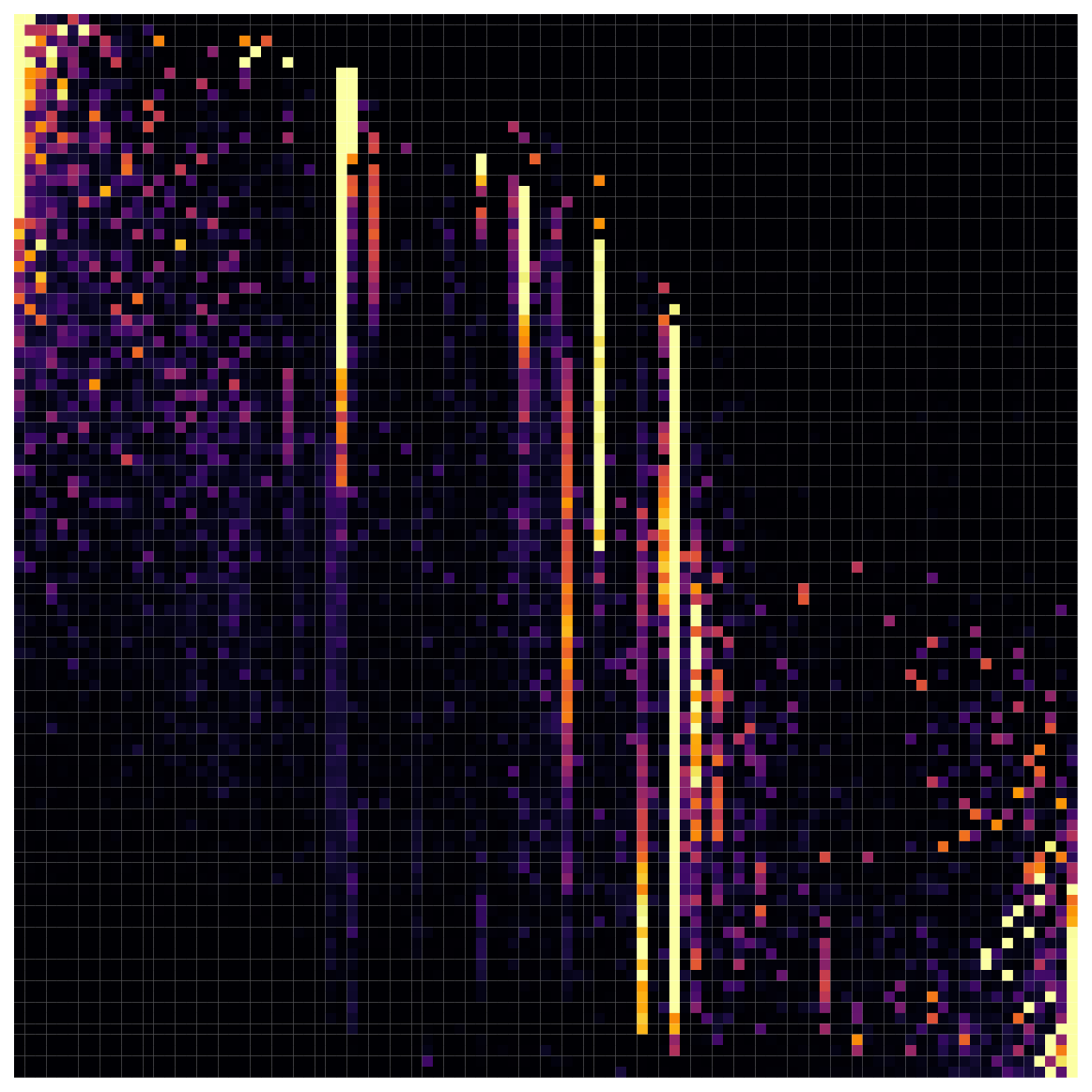}
\hfill
\includegraphics[width=0.49\textwidth]{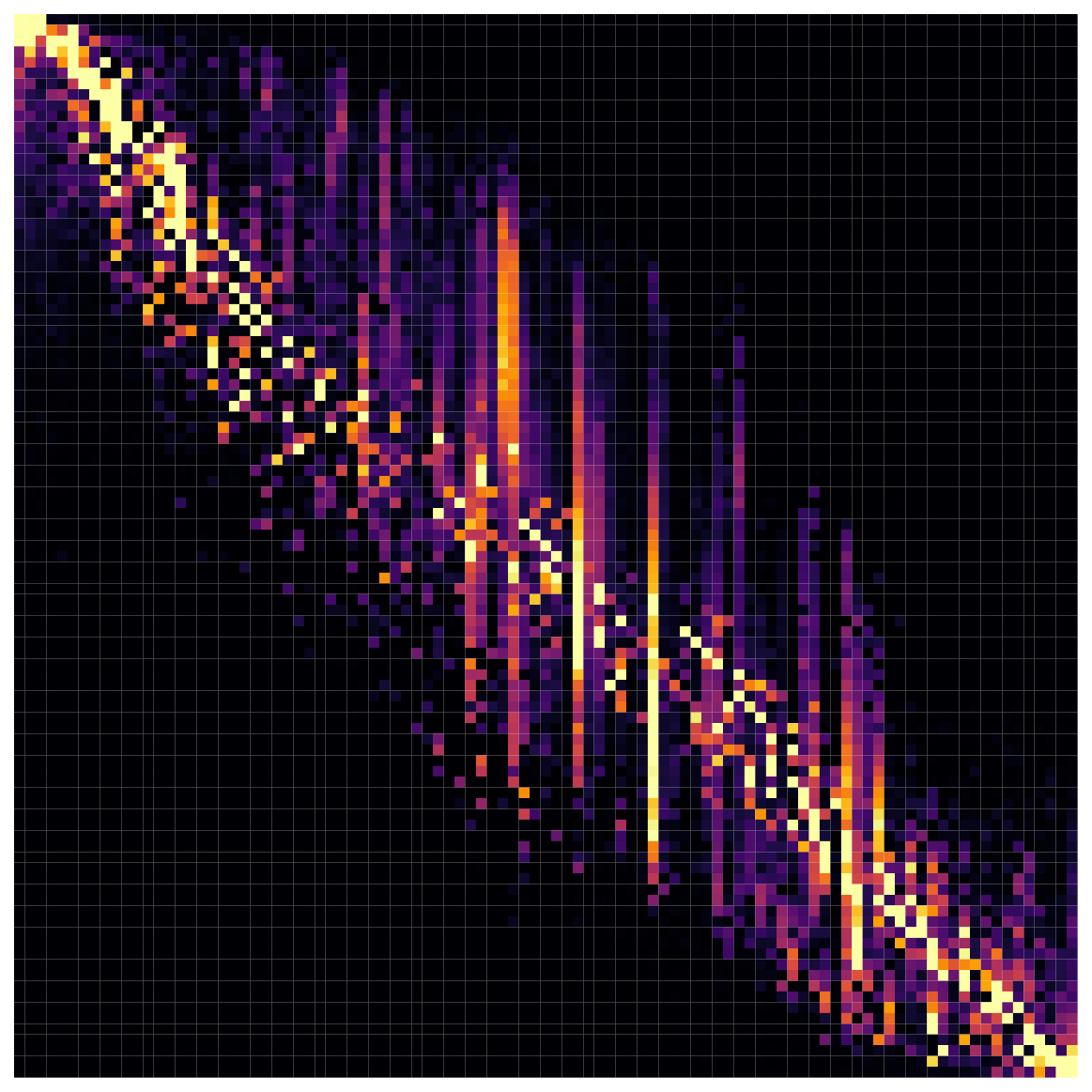}
\captionsetup{format=hang}
\caption{\it\small Heatmaps of aggregation weights fit by the local-fine (DQA)
  aggregator to the concrete data set (left) and the power data set (right). In
  the current fine setting, we have one $m \times m = 99 \times 99$ weight
  matrix per base model; the heatmaps have therefore been averaged over the
  $p=6$ base models. Their rows correspond to the output quantiles (for the 
  aggregator), and the columns to the input quantiles (from the base model). The
  right heatmap displays clear concentration around the diagonal: here each
  input quantile is on average the most reliable estimate used to define its 
  corresponding output quantile. This is much less true on the left, which also
  has more pronounced vertical ``streaks'': output quantiles use input quantiles
  from many neighboring quantile levels. The left heatmap also has notably more
  dispersion in the tails.}    
\label{fig:local_full_heatmap}
\end{figure}

Visualizing these attention maps---see Figures \ref{fig:local_medium_heatmap}
and \ref{fig:local_full_heatmap}---can help address natural questions, e.g., for
any particular $x$ and $\tau$, which base estimates are most useful? Or, how do
estimates at different quantile levels interact with each other to achieve
accuracy in the ensemble? While this conveys only \emph{qualitative}
information, we note that interpretations such as these would be far more
difficult to obtain for nonlinear aggregators like stacked ensembles
\citep{wolpert1992stacked}.

\subsection{New base model: deep quantile regression}

Finally, we remark that the optimization in \eqref{eq:linear_ensemble_opt_local}
can be used to itself define a standalone neural network quantile regressor,
which we refer to as \emph{deep quantile regression} or DQR. This is given by
taking $p=1$ in \eqref{eq:linear_ensemble_opt_local} with a trivial base model
that outputs \smash{$\hat{g}_1(x; \tau) = 1$}, for any $x$ and $\tau$. DQR can
be useful as a base model (to feed into an aggregation method like DQA), and
will be used as a point of comparison and/or illustration at various points in
what follows. For example, Figure \ref{fig:crossing_penalty} illustrates DQR on
a real data set.

\section{Quantile noncrossing}
\label{sec:noncrossing}

Here we discuss methods for enforcing quantile noncrossing in the predictions
produced by the aggregation methods from the last section. We combine and extend
the ideas introduced in Section \ref{sec:noncrossing_background}.

\subsection{Crossing penalty and quantile buffering}
\label{sec:cross_penalty}

As an alternative to enforcing hard noncrossing constraints during training, as
in \eqref{eq:multi_qr_constraints}, we consider a crossing penalty of the form:       
\begin{equation}
\label{eq:crossing_penalty}
\rho(g) = \sum_{x \in \nocrossx} \sum_{\tau<\tau'} \big( 
g(x; \tau) - g(x; \tau') + \delta_{\tau,\tau'} \big)_+,   
\end{equation}
where $a_+= \max\{a, 0\}$, and $\delta_{\tau,\tau'} \geq 0$, for $\tau < \tau'$ 
are constants (not depending on $x$) that encourage a buffer between pairs of
quantiles at different quantile levels. In the simplest case, we can reduce this
to just a single margin $\delta_{\tau, \tau'} = \delta$ across all quantile
levels, that we can tune as a hyperparameter (e.g., using a validation set). We
will cover a slightly more advanced strategy shortly, for fitting adaptive
margins which vary with quantile levels (as well as the training data at hand). 

\begin{figure}[htb]
\centering
\includegraphics[width=0.495\linewidth]{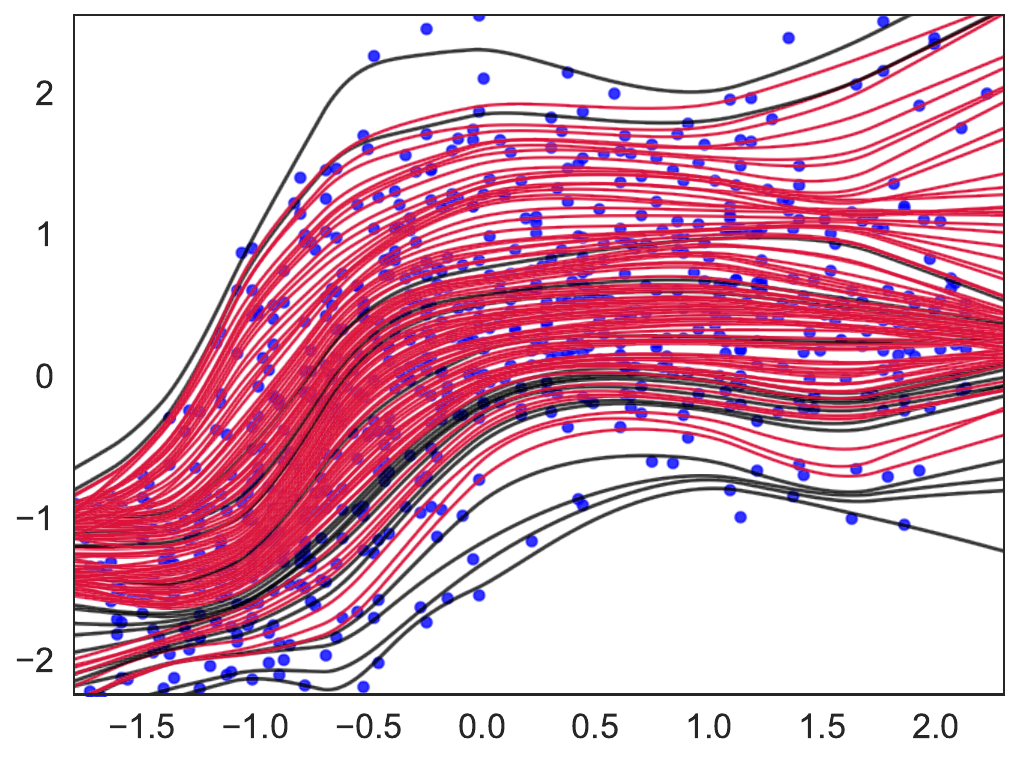}
\includegraphics[width=0.495\linewidth]{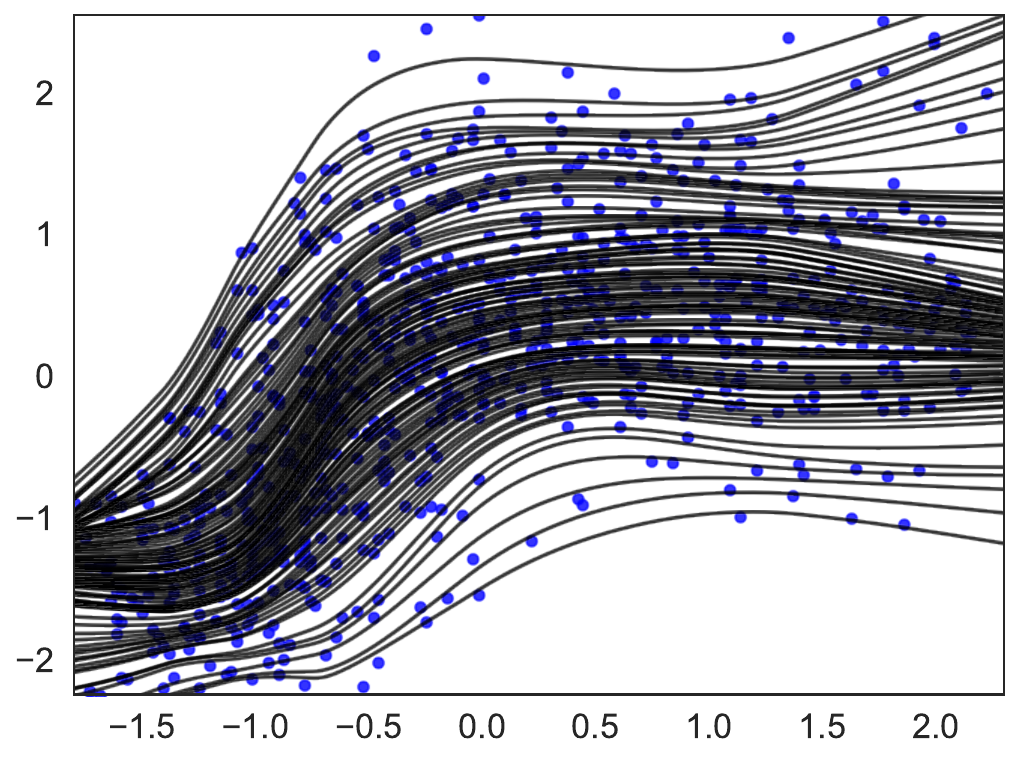}
\captionsetup{format=hang}
\caption{\it\small Conditional quantile estimates fit by a neural network
  quantile regressor (DQR) to the bone mineral data set, used in
  \citet{takeuchi2006nonparametric}, where $X$ is univariate. The left panel
  displays estimates that were fit without any crossing penalty or isotonization 
  operator, and the right panel displays estimates fit using the crossing
  penalty \eqref{eq:crossing_penalty} and post sorting. (This is the simplest 
  combination among all the methods that we consider to address quantile
  crossing.) Crossing quantile curves are highlighted in red.}    
\label{fig:crossing_penalty}
\end{figure}

The advantage of using a penalty \eqref{eq:crossing_penalty} over constraints is 
primarily computational: we can add it to the criterion defining the aggregation
model, and we can still use SGD for optimization. 
Note that the penalty is applied at the ensemble level, to \smash{$g =
  \hat{g}_w = \sum_{j=1}^p w_j \cdot \hat{g}_j$}; to be precise, the global
optimization \eqref{eq:linear_ensemble_opt} becomes     
\begin{equation}
\label{eq:linear_ensemble_opt_pen}
\begin{alignedat}{2}
&\minimize_w &&\frac{1}{n} \sum_{i=1}^n 
\sum_{\tau \in \Tcal} \pinball_\tau\bigg( Y_i - \sum_{j=1}^p w_j \cdot
\hat{g}_j^{-k(i)}(X_i; \tau) \bigg) + 
\lambda \, \rho\bigg( \sum_{j=1}^p w_j \cdot \hat{g}_j\bigg ) \\   
&\st \quad && Aw = 1, \quad w \geq 0, 
\end{alignedat}
\end{equation} 
where $\lambda \geq 0$ is a hyperparameter to be tuned. The modification of the
local optimization \eqref{eq:linear_ensemble_opt_local} to include the penalty
term is similar. Figure \ref{fig:crossing_penalty} gives an illustration of the 
crossing penalty in action.

\subsubsection{Adaptive margins}

One of the problems with choosing a fixed margin $\delta$ (of using the
reduction $\delta_{\tau,\tau'} = \delta$ for all $\tau,\tau'$) is that it may
not accurately reflect the scale or shape of the quantile gaps in the data at
hand. We propose the following heuristic for tuning margin parameters
$\delta_{\tau, \tau'}$, over all pairs of distinct levels $\tau < \tau'$.  
We begin with any pilot estimator of the conditional mean or median of $Y|X$,
call it \smash{$\hat{g}_0$}, and form residuals $R_i = Y_i - \hat{g}_0(X_i)$,
$i=1,\ldots,n$. The closer these residuals are to the true error distribution, 
of $Y - \E(Y|X)$ (or similar for the median), the better. Thus we typically
want to use cross-validation residuals for this pilot step. We then take, for
each $\tau < \tau'$, the margin to be   
\begin{equation}
\label{eq:adapt_margin}
\delta_{\tau,\tau'} = \delta_0 \big( Q_{\tau'}(\{R_i\}_{i=1}^n) -
Q_{\tau}(\{R_i\}_{i=1}^n) \big)_+.
\end{equation}
where $Q_\tau$ gives the empirical level $\tau$ quantile of its argument. The
motivation behind the proposal in \eqref{eq:adapt_margin} is that
\smash{$Q_{\tau'}(\{R_i\}_{i=1}^n) - Q_{\tau}(\{R_i\}_{i=1}^n)$} is itself the
gap between quantile estimates from a basic but properly (monotonically)
ordered quantile regressor, namely, that defined by:
\[
\hat{g}_0(x; \tau) = \hat{g}_0(x) + Q_{\tau}(\{R_i\}_{i=1}^n), \quad x \in
\Xcal, \, \tau \in \Tcal.
\]
We note that $\delta_0 \geq 0$ in \eqref{eq:adapt_margin} is a single
hyperparameter to be tuned. In our experiments, we take the pilot estimator 
to be the simple average of base model predictions at the median,
\smash{$\hat{g}_0(x) = \frac{1}{p} \sum_{j=1}^p   \hat{g}_j(x; 0.5)$}, and we
use the same cross-validation scheme for the base models as in Section
\ref{sec:oos_base} to produce out-of-fold residuals $R_i$, $i=1,\ldots,n$
around \smash{$\hat{g}_0$}, that are used in \eqref{eq:adapt_margin}.     

\subsection{Isotonization: post hoc and end-to-end}
\label{sec:iso_post_grad}

Next we consider different isotonization operators that can be used in
conjunction with the crossing penalty in \eqref{eq:crossing_penalty}. This is
important because it will guarantee quantile noncrossing for the ensemble
predictions at all points $x \in \Xcal$, as in \eqref{eq:noncrossing}, whereas
the crossing penalty alone is not sufficient to ensure this. The added benefit
of this approach is that we are able to satisfy the constraint at out-of-sample
points without having to invest unreasonable amounts of function complexity into 
the model, simply by applying isotonization as a form of prior knowledge (see  
\citealt{le2006simpler} for a detailed general discussion of this aspect).   

Generally, an isotonization operator $\Scal$ can be used in one of two ways. The
first is \emph{post hoc}, as explained in Section
\ref{sec:noncrossing_background}, where we simply apply it to quantile
predictions after the ensemble \smash{$\hat{g} = \sum_{j=1}^p \hat{w}_j \cdot
\hat{g}_j$} has been trained, as in \eqref{eq:post_hoc}. The second way is
\emph{end-to-end}, where $\Scal$ gets used a layer in training, and we now
solve, instead of \eqref{eq:linear_ensemble_opt_pen}, 
\begin{equation}
\label{eq:linear_ensemble_opt_pen_iso}
\begin{alignedat}{2}
&\minimize_w &&\frac{1}{n} \sum_{i=1}^n 
\sum_{\tau \in \Tcal} \pinball_\tau\Bigg( Y_i - \bigg[ \Scal\bigg( \sum_{j=1}^p 
w_j \cdot  \hat{g}_j^{-k(i)}(X_i) \bigg) \bigg]_\tau \Bigg) + 
\lambda \, \rho\bigg( \sum_{j=1}^p w_j \cdot \hat{g}_j\bigg ) \\   
&\st \quad && Aw = 1, \quad w \geq 0,
\end{alignedat}
\end{equation} 
where we denote by $[\Scal(v)]_\tau$ the element of $\Scal(v)$ corresponding to
a quantile level $\tau$. The modification for the local aggregation problem is  
similar. For predictions, we still post-apply $\Scal$, as in
\eqref{eq:post_hoc}.         

One might imagine that the end-to-end approach can help predictive accuracy,
since in training we are leveraging the knowledge of what we will be doing at
prediction time (applying $\Scal$). Furthermore, by treating $\Scal$ as a
differentiable layer, we can still seamlessly apply SGD for optimization
(end-to-end training). Note that sorting \eqref{eq:sort} and isotonic projection
\eqref{eq:iso_proj} are almost everywhere differentiable (they are in fact
almost everywhere linear maps). We will propose an additional isotonization
operator shortly that is also almost everywhere differentiable, and particularly
convenient to implement in modern deep learning toolkits. 

To be clear, the end-to-end approach is not free of downsides; while sorting and
isotonic projection are guaranteed to improve WIS when applied post hoc
(Proposition \ref{prop:post_hoc_wis}), the same cannot be said when they are
used end-to-end. Thus we give up on a formal guarantee, to potentially gain 
accuracy in practice.  

\subsubsection{Min-max sweep}

In addition to sorting \eqref{eq:sort} and isotonic projection
\eqref{eq:iso_proj}, we propose and investigate a simple isotonization operator
that we call the \emph{min-max sweep}. It works as follows: starting at the
median, it makes two outward sweeps: one upward and one downward, where it
replaces each value with a cumulative max (upward sweep) or cumulative min
(downward sweep). To be more precise, if we have sorted quantile levels $\tau_1
< \cdots < \tau_m$ with $\tau_{m_0} = 0.5$, and corresponding values $v_1,
\ldots, v_m$, then we define the \emph{min-max sweep} operator according to:  
\begin{equation}
\label{eq:min_max_sweep}
\mms(v)_k = 
\begin{cases}
v_k & k = m_0 \\
\max\{v_k, \, \mms(v)_{k-1}\} & k = m_0+1,\ldots,m \\
\min\{v_k, \, \mms(v)_{k+1}\} & k = m_0-1,\ldots,1.
\end{cases}
\end{equation}
Like sorting and isotonic projection, the min-max sweep is almost everywhere a
linear map, and thus almost everywhere differentiable. The primary motivation
for it is that it can be implemented highly efficiently in modern deep learning
frameworks via cumulative max and cumulative min functions---these are
vectorized operations that can be efficiently applied in parallel over a
mini-batch, in each iteration of SGD (when it is used in an end-to-end
manner). Unlike sorting and isotonic projection, however, it is not guaranteed 
to improve WIS (Proposition \ref{prop:post_hoc_wis}) when applied in post.

\section{Conformal calibration}
\label{sec:conformal}

Informally, a conditional quantile estimator is said to be \emph{calibrated}
provided that its quantile estimates have the appropriated nominal one-sided
coverage, for example, an estimated 90\% quantile covers the target from above
90\% of the time, and the same for the other quantile levels being
estimated. Equivalently, we can view this in terms of central prediction
intervals: the interval whose endpoints are given by the estimated 5\% and 95\%
quantiles covers the target 90\% of the time. Formally, let \smash{$\hat{g}$} be
an estimator trained on samples $(X_i,Y_i)$, $i=1,\ldots,n$, with
\smash{$\hat{g}(x; \tau)$} denoting the estimated level $\tau$ quantile of
$Y|X=x$. Assume that the set of quantile levels $\Tcal$ being estimated is
symmetric around 0.5, so we can write it as \smash{$\Tcal = \cup_{\alpha \in
\Acal} \{ \alpha/2, 1-\alpha/2 \}$}. Then \smash{$\hat{g}$} is said to be
calibrated, provided that for every $\alpha \in \Acal$, we have
\begin{equation}
\label{eq:calibration}
\P\Big(Y_{n+1} \in \Big[\hat{g}(X_{n+1}; {\alpha}/{2}), \, 
\hat{g}(X_{n+1}; 1-{\alpha}/{2}) \Big]\Big) = 1-\alpha,
\end{equation}
where $(X_{n+1},Y_{n+1})$ is a test sample, assumed to be i.i.d.\ with the
training samples $(X_i,Y_i)$, $i=1,\ldots,n$. To be precise, the notion of
calibration that we consider in \eqref{eq:calibration} is \emph{marginal} over
everything: the training set and the test sample. We remark that a stronger
notion of calibration, which may often be desirable in practice, is calibration  
\emph{conditional} on the test feature value:
\begin{equation}
\label{eq:calibration_cond}
\P\Big(Y_{n+1} \in \Big[\hat{g}(X_{n+1}; {\alpha}/{2}), \, 
\hat{g}(X_{n+1}; 1-{\alpha}/{2}) \Big] \,\Big|\, X_{n+1}=x \Big) =
1-\alpha, \quad x \in \Xcal. 
\end{equation}
Conditional calibration, as in \eqref{eq:calibration_cond}, is actually
impossible to achieve in a distribution-free sense; see
\citet{lei2014distribution, vovk2012conditional, barber2021limits} for precise
statements and developments. Meanwhile, marginal calibration
\eqref{eq:calibration} is possible to achieve using \emph{conformal prediction}
methodology, without assuming anything about the joint distribution of
$(X,Y)$. The definitive reference on conformal prediction is the book by
\citet{vovk2005algorithmic}; see also \citet{lei2018distribution}, which helped
popularize conformal methods in the statistics and machine learning
communities. The conformal literature is by now somewhat vast, but we will only
need to discuss a few parts of it that are most relevant to calibration of
conditional quantile estimators.

In particular, we very briefly review \emph{conformalized quantile regression}
or CQR by \citet{romano2019conformalized} and CV+ by
 \citet{barber2021predictive}. We then discuss how these can be efficiently
 applied in combination to any of the quantile aggregators studied in this
 paper.   

\subsection{CQR}

We first describe CQR in a split-sample setting. Suppose that we have reserved a
subset indexed by $\Ical_1 \subseteq \{1,\ldots,n\}$ as the \emph{proper 
  training set}, and $\Ical_2 = \{1,\ldots,n\} \setminus \Ical_1$ as the
\emph{calibration set}. (Extending this to a cross-validation setting, via the
CV+ method, is discussed next.) In this split-sample setting, CQR from
\citet{romano2019conformalized} can be explained as follows.  

\begin{enumerate}
\item First, fit the quantile estimator \smash{$\hat{g}^{\Ical_1}$} on the
  proper training set $\{(X_i,Y_i) : i \in \Ical_1\}$. 
\item Then, for each $\alpha \in \Acal$, compute lower and upper residuals on
  the calibration set,
  \[
  R^{-}_{i,\alpha} = \hat{g}^{\Ical_1}(X_i; \alpha/2) - Y_i
  \quad \text{and} \quad
  R^{+}_{i,\alpha} = Y_i - \hat{g}^{\Ical_1}(X_i; 1-\alpha/2), 
  \quad i \in \Ical_2.  
  \]
\item Finally, adjust (or \emph{conformalize}) the original estimates based on 
  residual quantiles, yielding for $\alpha \in \Acal$ and $x \in \Xcal$,   
  \begin{equation}  
  \begin{aligned}
  \label{eq:cqr_split}
  \tilde{g}(x; \alpha/2) &= \hat{g}^{\Ical_1}(x; \alpha/2) -
  \tilde{Q}_{1-\alpha/2}\big( \{R^{-}_{i,\alpha} : i \in \Ical_2\} \big), \\
  \tilde{g}(x; 1-\alpha/2) &= \hat{g}^{\Ical_1}(x; 1-\alpha/2) +
  \tilde{Q}_{1-\alpha/2}\big( \{R^{+}_{i,\alpha} : i \in \Ical_2\} \big), 
  \end{aligned}
  \end{equation}
  where \smash{$\tilde{Q}_\tau$} is a slightly modified empirical quantile
  function, giving the empirical level $\lceil \tau(n_2+1) \rceil / n_2$
  quantile of its argument, with $n_2 = |\Ical_2|$.    
\end{enumerate}

A cornerstone result in conformal prediction theory (see Theorems 1 and 2 in
\citet{romano2019conformalized} for the application of this result to CQR) says
that for any estimator, its conformalized version in \eqref{eq:cqr_split} has
the finite-sample marginal coverage guarantee 
\begin{equation}
\label{eq:cqr_split_cov}
\P\Big(Y_{n+1} \in \Big[\tilde{g}(X_{n+1}; {\alpha}/{2}), \, 
\tilde{g}(X_{n+1}; 1-{\alpha}/{2}) \Big]\Big) \geq 1-\alpha.
\end{equation}
In fact, the coverage of the level $1-\alpha$ central prediction interval is
also upper bounded by $1-\alpha+1/(n_2+1)$, provided that the residuals are
continuously distributed (i.e., there are almost surely no ties).

\subsection{CV+}

Now we describe how to extend CQR to a cross-validation setting. Let $\{ I_k
\}_{k=1}^K$ denote a partition of $\{1,\ldots,n\}$ into disjoint folds. We can
map steps 1--3 used to produce the CQR correction in \eqref{eq:cqr_split} onto
the CV+ framework of \citet{barber2021predictive}, as follows.  

\begin{enumerate}
\item First, for each fold $k=1,\ldots,K$, fit the quantile estimator
  \smash{$\hat{g}^{-k}$} on all data points but those in the $k\th$ fold,
  $\{(X_i,Y_i) : i \notin \Ical_k\}$.  
\item Then, for each $i=1,\ldots,n$ and each $\alpha \in \Acal$, compute lower  
  and upper residuals on the calibration fold $k(i)$ (where $k(i)$ denotes the
  index of the fold containing the $i\th$ data point),
  \[
  R^{-}_{i,\alpha} = \hat{g}^{-k(i)}(X_i; \alpha/2) - Y_i
  \quad \text{and} \quad
  R^{+}_{i,\alpha} = Y_i - \hat{g}^{-k(i)}(X_i; 1-\alpha/2),
  \quad i \in \Ical_{k(i)}.
  \]
\item Finally, adjust (or conformalize) the original estimates based on residual
  quantiles, yielding for $\alpha \in \Acal$ and $x \in \Xcal$,    
  \begin{equation}  
  \begin{aligned}
  \label{eq:cqr_cv+}
  \tilde{g}(x; \alpha/2) &= \tilde{Q}^{-}_{1-\alpha/2} \Big( \big\{
    \hat{g}^{-k(i)}(x; \alpha/2) - R^{-}_{i,\alpha} : i \in k(i) \big\} \Big), \\  
  \tilde{g}(x; 1-\alpha/2) &= \tilde{Q}^{+}_{1-\alpha/2} \Big( \big\{
    \hat{g}^{-k(i)}(x; 1-\alpha/2) + R^{+}_{i,\alpha} : i \in k(i) \big\} \Big),
  \end{aligned}
  \end{equation}
  where now \smash{$\tilde{Q}^{-}_\tau, \tilde{Q}^+_\tau$} are modified
  lower and upper level $\tau$ empirical quantiles: for any set $Z$, we define
  \smash{$\tilde{Q}^+_\tau(Z)$} to be the level $\lceil \tau(|Z|+1) \rceil /
  |Z|$ empirical quantile of $Z$, and \smash{$\tilde{Q}^-_\tau(Z) =
    -\tilde{Q}^+(-Z)$}.    
\end{enumerate}

According to results from \citet{barber2021predictive} and \citet{vovk2018cross}
(see Theorem 4 in \citet{barber2021predictive} for a summary), for any
estimator, its conformalized version in \eqref{eq:cqr_cv+} has the finite-sample
marginal coverage guarantee  
\begin{equation}
\label{eq:cqr_cv+_cov}
\begin{aligned}
\P\Big(Y_{n+1} \in \Big[\tilde{g}(X_{n+1}; {\alpha}/{2}), \, 
\tilde{g}(X_{n+1}; 1-{\alpha}/{2}) \Big]\Big) &\geq 1-2\alpha - 
\min\bigg\{\frac{2(1-1/K)}{n/K+1}, \, \frac{1-K/n}{K+1}\bigg\} \\
&\geq 1- 2\alpha - \sqrt{2/n}.
\end{aligned}
\end{equation}
We can see that the guarantee from CV+ in \eqref{eq:cqr_cv+_cov} is weaker than
that in \eqref{eq:cqr_split_cov} from sample-splitting, though CV+ has the
advantage that it utilizes each data point for \emph{both} training and
calibration purposes, and can often deliver shorter conformalized prediction
intervals as a result. Moreover, \citet{barber2021predictive} argue that for
stable estimation procedures, the empirical coverage from CV+ is closer to
$1-\alpha$ (and provide some theory to support this as well).

\subsection{Nested implementation for quantile aggregators}

We discuss the application of CQR and CV+ to calibrate the quantile aggregation
models developed in Section \ref{sec:methods} and \ref{sec:noncrossing}. An
additional level of complexity in this application arises because the quantile
aggregation model is itself built using cross-validation: recall, as discussed
in Section \ref{sec:oos_base}, that the aggregation weights are trained using 
out-of-sample (out-of-fold) predictions from the base models. Thus, application
of CV+ here requires a \emph{nested} cross-validation scheme, where in the inner 
loop, the aggregator is trained using cross-validation (for the base model
predictions), and in the outer loop, the calibration residuals are computed for
the ultimate conformalization step. 

In order to make this as computationally efficient as possible, we introduce a
particular nesting scheme that reduces the number of times base models are
trained by roughly a factor of 2. We first pick a number of folds $K$ for the
outer loop, and define folds $\Ical_k$, $k=1,\ldots,K$. Then for each
$k=1,\ldots,K$, we fit the quantile aggregation model \smash{$\hat{g}_w^{-k}$}
on $\{(X_i,Y_i): i \notin \Ical_k\}$ (using any one of the approaches described
in Sections \ref{sec:methods} and \ref{sec:noncrossing}), with the key idea
being how we implement the inner cross-validation loop to train the base
models: we use folds $\Ical_\ell$, $\ell \not= k$ for this inner loop, so that
we can later avoid having to refit the base models when the roles of $k$ and
$\ell$ are reversed. To see more clearly why this is the case, we describe the 
procedure in more detail below.

\begin{enumerate}
\item For $k=1,\ldots,K$:
\begin{enumerate}[label=\alph*.]
\item For $\ell \not= k$, train each base model \smash{$\hat{g}_j^{-k,\ell}$} on
  $\{(X_i,Y_i): i \notin \Ical_k \cup \Ical_\ell\}$.
\item Train the aggregation weights $w$ on $\{(X_i,Y_i): i \notin \Ical_k\}$ 
  (using the base model predictions \smash{$\hat{g}_j^{-k,\ell}(X_i)$} from Step
  a), to produce the aggregator \smash{$\hat{g}_w^{-k}$}. 
\item Compute lower and upper calibration residuals of \smash{$\hat{g}_w^{-k}$}
  on $\{(X_i,Y_i) : i \in \Ical_k\}$, as in Step 2 of CV+.
\end{enumerate}
\item Conformalize original estimates using residual quantiles, as in Step 3 of
  CV+. 
\end{enumerate}

The computational savings comes from observing that \smash{$\hat{g}_j^{-k,\ell}
= \hat{g}_j^{-\ell,k}$}: the base model we train when $k$ is the calibration
fold and $\ell$ is the validation fold is the same as that we would train when
the roles of $k$ and $\ell$ are reversed. Therefore we only need to train each
base model in Step 1a, across all iterations of the outer and inner loops, a
total of $K(K-1)/2$ times, as opposed to $K(K-1)$ times if we were ignorant to
this fact (or $K^2$ times, the result of a more typical nested $K$-fold
cross-validation scheme).  

\section{Empirical comparisons}
\label{sec:empirical}

We provide a broad empirical comparison of our proposed quantile aggregation 
procedures as well as many different established regression and aggregation
methods. We examine 34 data sets in total: 8 from the UCI Machine Learning
Repository \citep{dua2017uci} and 26 from the AutoML Benchmark for Regression
from the Open ML Repository \citep{vanschoren2013open}. These have served as
popular benchmark repositories in probabilistic deep learning and uncertainty
estimation, see, e.g., \citet{lakshminarayanan2017uncertainty,
  mukhoti2018importance, jain2020maximizing, fakoor2020fast}. 
To the best of our knowledge, what follows is the most comprehensive evaluation
of quantile regression/aggregation methods published to date.  

\subsection{Training, validation, and testing setup}

For each of the 34 data sets studied, we average all results over 5 random
train-validation-test splits, of relative size 72\% (train), 18\% (validation),
and 10\% (test). We adopt the following workflow for each data set.    

\begin{enumerate}
\item Standardize the feature variables and response variable on the combined
  training and validation sets, to have zero mean and unit standard deviation. 
\item Fit the base models on the training set.
\item Find the best hyperparameter configuration for each base model by
  minimizing weighted interval score (WIS) on the validation set. Fix these
  henceforth. 
\item Divide the training set into 5 random folds. Record out-of-fold (OOF)
  predictions for each base model; that is, for fold $k$, we fit each base model
  \smash{$\hat{g}_j^{-k}$} by training on data from all folds except $k$, and
  use it to produce \smash{$\hat{g}_j^{-k}(X_i)$} for each $i$ such that $k(i) =
  k$.  
\item Fit the aggregation models on the training set, using the OOF base
  predictions from the last step.  
\item Find the best hyperparameter configuration for each aggregation model by
  again minimizing validation WIS. Fix these henceforth.  
\item Report the test WIS for each aggregation model (making sure to account for
  the initial standardization step, in the test set predictions).        
\end{enumerate} 

All base and aggregation models are fit using $m=99$ levels, $\Tcal = \{0.01,  
0.02, \ldots, 0.99\}$; the specific base and aggregation models we consider are
described below. The initial standardization step is convenient for
hyperparameter tuning. For more details on hyperparameter tuning, see 
Appendix \ref{app:base_models} and \ref{app:aggr_models}.

\subsection{Base models}

We consider $p=6$ quantile regression models as base models for the
aggregators, described below, along with the abbreviated names we give them
henceforth. The first three are neural network models with different
architectures and objectives. More details are given in Appendix
\ref{app:base_models}. 

\begin{enumerate}
\item Conditional Gaussian network (CGN,
  \citealp{lakshminarayanan2017uncertainty}).
\item Simultaneous quantile regression (SQR, \citealp{tagasovska2019single});  
\item Deep quantile regression (DQR), which falls naturally out of our
  framework, as explained in Section \ref{sec:local_aggr}.
\item Quantile random forest (RandomForest, \citealp{meinshausen2006quantile}).
\item Extremely randomized trees (ExtraTrees, \citealp{geurts2006extremely}).
\item Quantile gradient boosting (LightGBM, \citealp{ke2017gbm}).
\end{enumerate}

\subsection{Aggregation models}

We study 4 aggregation models outside of the ones defined in our paper,
described below, along with their abbreviated names.  More details are given in
Appendix \ref{app:aggr_models}. 

\begin{enumerate}
\item Average: a straight per-quantile average of base model predictions. 
\item Median: a straight per-quantile median of base model predictions. 
\item Quantile regression averaging (QRA, \citealp{nowotarski2015computing}).
\item Factor QRA (FQRA, \citealp{maciejowska2016probabilistic}).
\end{enumerate} 

The rest of this section proceeds as follows. We first discuss the performance
of the local-fine aggregator from our framework in Section \ref{sec:methods},
called DQA, to other methods (all base models, and the other aggregators defined
outside of this paper). In the subsections that follow, we move on to a finer
analysis, comparing the different weighted ensembling strategies to each other,
comparing the different isotonization approaches to each other, and evaluating
the gains in calibration made possible by conformal prediction methodology.     

\subsection{Continuation of Figure \ref{fig:intro}}
\label{sec:continuation_fig1}

Recall that in Figure \ref{fig:intro}, we already described some results from our 
empirical study, comparing DQA to all other methods. To be precise, for
DQA---in Figure \ref{fig:intro}, here, and henceforth---we use the simplest
strategy to ensure quantile noncrossing among the options discussed in Section
\ref{sec:noncrossing}: the crossing penalty along with post sorting
(CrossPenalty + PostSort in the notation that we will introduce shortly). To
give an aggregate view of the same results, Figure \ref{fig:intro_bar} shows the
average relative WIS over all data sets. Relative WIS is the ratio of the WIS of
given method to that of DQA, hence 1 indicates equal performance to DQA, and
larger numbers indicate worse performance. (Standard error bars are also drawn.)
These results clearly show the favorable average-case performance of DQA, to
complement the per-data-set view in Figure \ref{fig:intro}. We can also see that
QRA and FQRA are runners up in terms of average performance; while they are
fairly simple aggregators, based on quantile linear regression, they use inputs
here the predictions from flexible base models. Going back to the results in
Figure \ref{fig:intro}, it is worth noting that there are data sets for which
QRA and FQRA perform clearly worse than DQA (up to 50\% worse, a relative
factor of 1.5), but none where they perform better. In Appendix
\ref{app:intro_revisit}, we provide an alternative visualization of these same 
results (differences in WIS rather than ratios of WIS).

\subsection{Comparing ensembling strategies} 

\begin{figure}[p]
\centering
\includegraphics[width=0.93\textwidth]{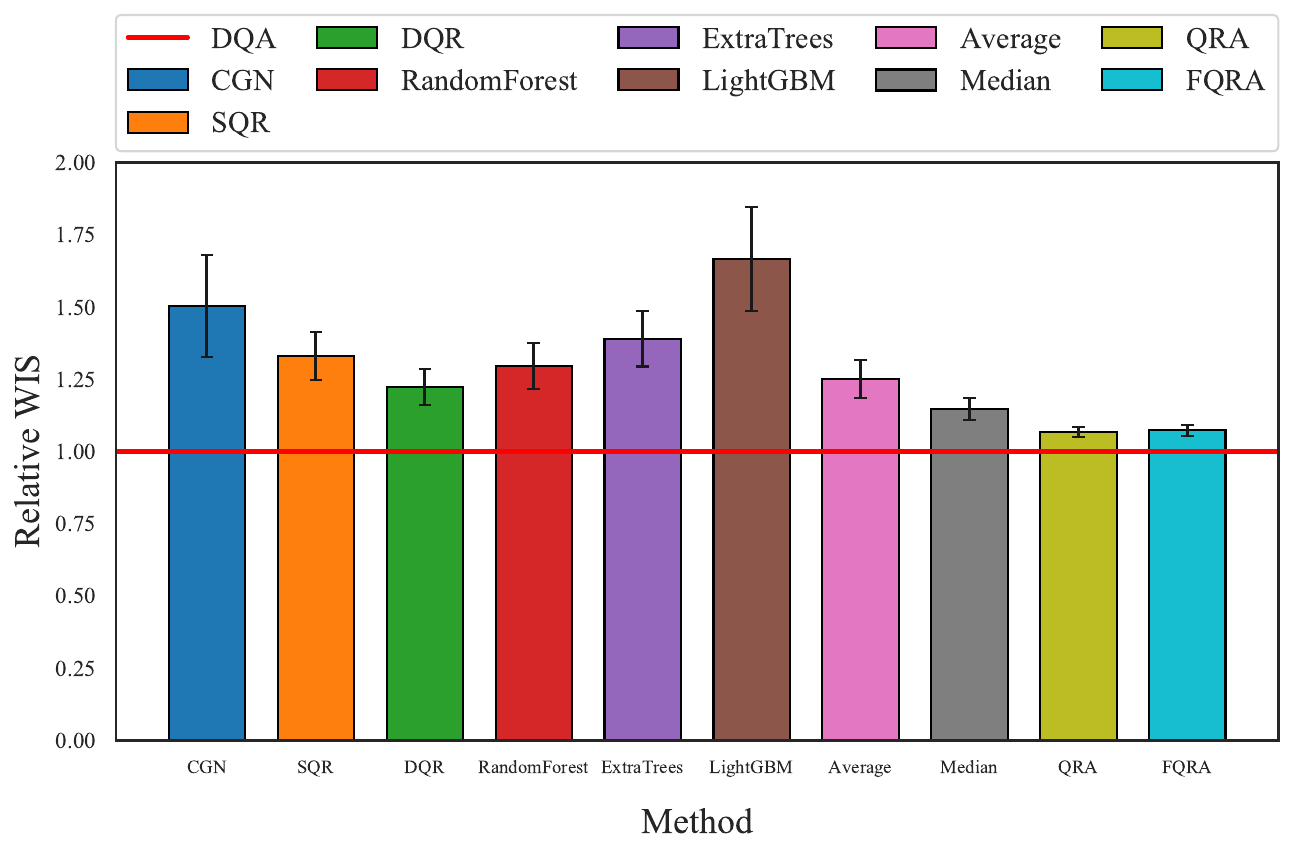}
\captionsetup{format=hang}
\caption{\it\small Average relative WIS over all data sets for deep quantile
  aggregation (DQA) and various quantile regression methods. The per-data-set
  results are given in Figure \ref{fig:intro}. Numbers larger than 1 indicate a
  worse average performance than DQA. We see that DQA performs the best overall,
  and QRA and FQRA are somewhat close runners up.} 
\label{fig:intro_bar}

\bigskip\bigskip
\includegraphics[width=\textwidth]{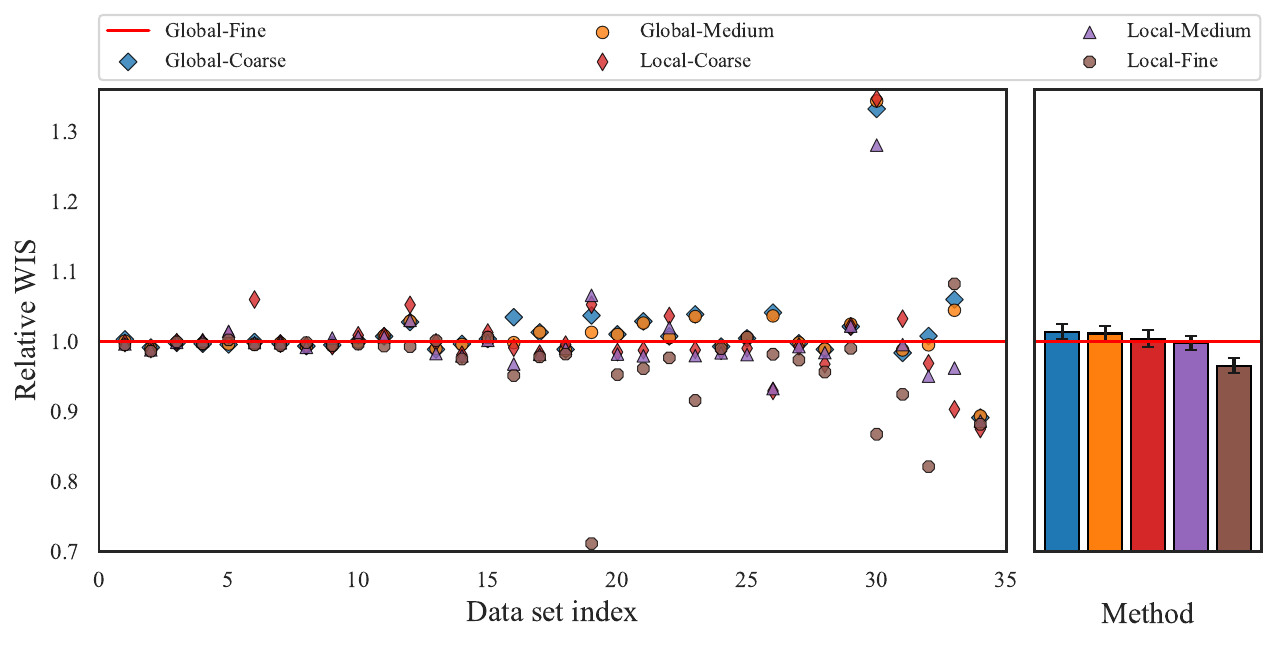}
\captionsetup{format=hang}
\caption{\it\small Comparing the relative WIS of various ensembling strategies
  to global-fine. The left panel shows the relative WIS per data set (ordered by  
  increasing PVE), and the right panel shows the average relative WIS over all 
  data sets. In general, the more flexible aggregation strategies tend to
  perform better, and local-fine (DQA) performs the best overall.}  
\label{fig:ensembling_strategies}
\end{figure}

Now we compare different ensembling strategies of varying flexibility, as
discussed in Section \ref{sec:methods}. Recall that we have two axes: global or
local weighting; and coarse, medium, or fine parameterization. This gives a
total of 6 approaches, which we denote by global-coarse, global-medium, and so
on. Figure \ref{fig:ensembling_strategies} displays the relative WIS of each 
method to global-fine per data set (left panel), and the average relative WIS
over all data sets (right panel). The data sets here and henceforth are indexed
by increasing PVE, as in Figure \ref{fig:intro}. We note that global-fine
is like a ``middle man'' in terms of its flexibility, among the 6 strategies
considered. The overall trend in Figure \ref{fig:ensembling_strategies} is that
greater flexibility leads to better performance, especially for larger PVE
values.     

\subsection{Comparing isotonization approaches}

\begin{figure}[!t]
\includegraphics[width=\textwidth]{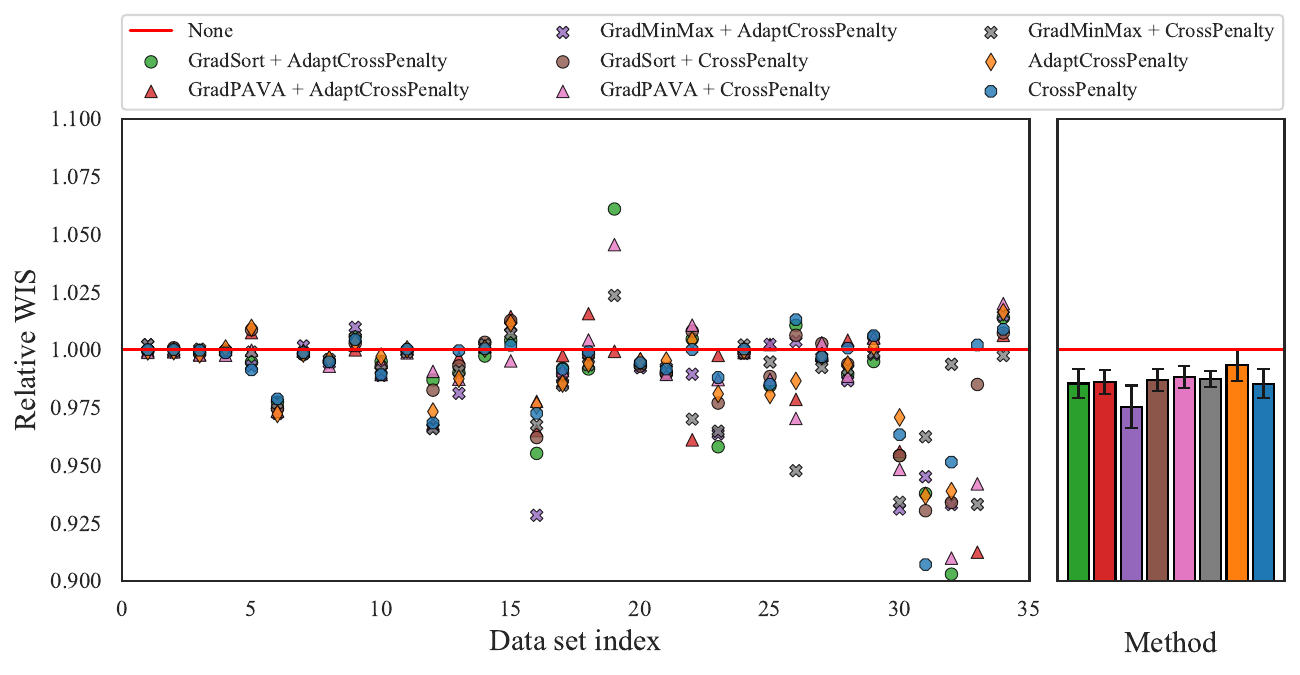}
\captionsetup{format=hang}
\caption{\it\small Comparing the relative WIS of various isotonization
  strategies on top of pure DQA, with no penalty and no isotonization operator  
  (``none''). The display is as in Figure \ref{fig:ensembling_strategies}  
  (individual results per data set on left panel, averaged results on right
  panel). We can see that all isotonization methods offer consistent but small 
  improvements on ``none'', with AdaptCrossPenalty + GradMinMax achieving the 
  best average-case performance.}
\label{fig:iso_strategies}
\end{figure}

We compare the various isotonization approaches discussed in Section
\ref{sec:noncrossing}, applied to DQA. In Figure \ref{fig:iso_strategies}, the
legend label ``none'' refers to DQA without any penalty and any isotonization
operator; all WIS scores are computed relative to this method. CrossPenalty and
AdaptCrossPenalty, respectively, denote the crossing penalty and adaptive cross
penalty from Section \ref{sec:cross_penalty}. PostSort and GradSort,
respectively, denote the sorting operator applied in post-processing and in
training (see \eqref{eq:linear_ensemble_opt_pen_iso} for how this works in the
global models), as described in Section \ref{sec:iso_post_grad}. We similarly
use PostPAVA, GradPAVA, and so on. Lastly, we use ``+'' to denote the
combination of a penalty and isotonization operator, as in CrossPenalty +
PostSort. The main takeaway from Figure \ref{fig:iso_strategies} is that all of
the considered methods improve upon ``None''. However, it should be noted that
these are second-order improvements in relative WIS compare to the much larger
first-order improvements in Figures \ref{fig:intro} and \ref{fig:intro_bar}, of
DQA over base models and other aggregators---compare the y-axis range in 
Figure \ref{fig:iso_strategies} to that in Figures \ref{fig:intro} and 
\ref{fig:intro_bar}. 

A second takeaway from Figure \ref{fig:iso_strategies} might be to generally
recommend AdaptCrossPenalty + GradMinMax for future aggregation problems, which
had the best average performance across our empirical suite. That said, given
the small relative improvement this offers, we generally stick to using the
simpler CrossPenalty + PostSort in the current paper.

\begin{figure}[!t]
\centering
\includegraphics[width=0.75\textwidth]{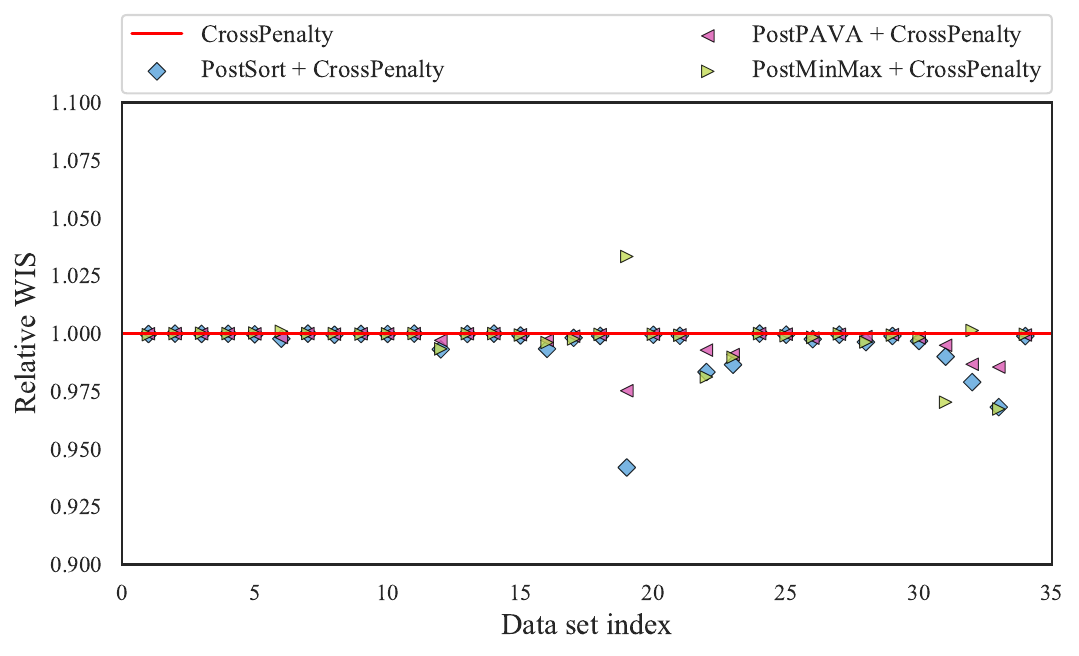}
\includegraphics[width=0.75\textwidth]{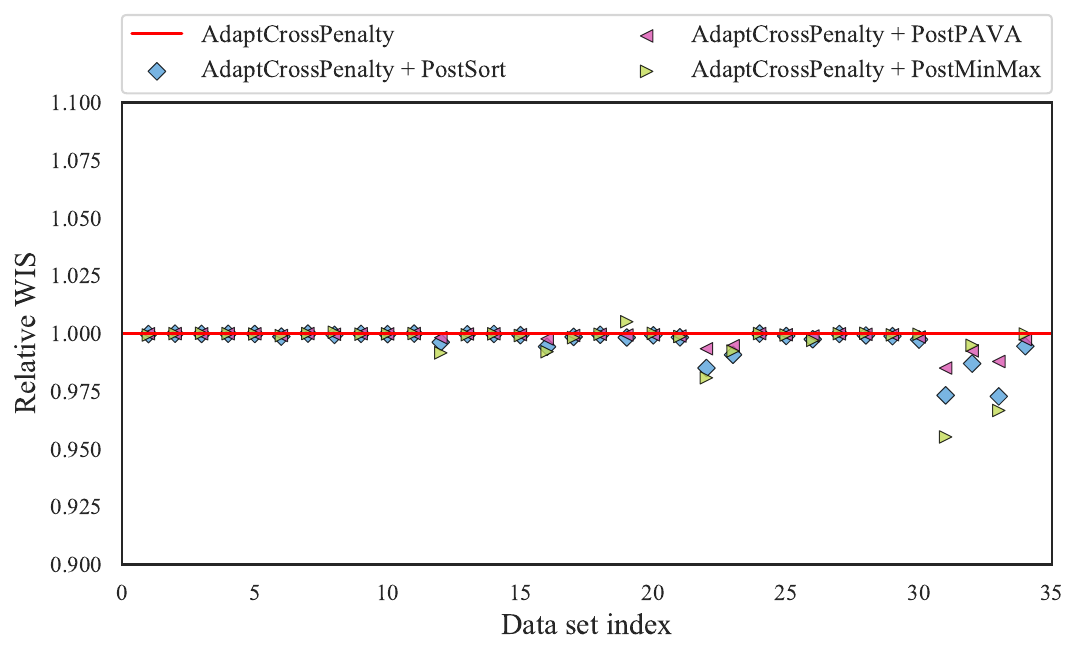}
\captionsetup{format=hang}
\caption{\it\small Comparing post isotonization strategies applied on top of
  CrossPenalty (shown in the top figure), and AdaptCrossPenalty (shown in the bottom figure). As we can see in either figure, PostSort and PostPAVA can only improve WIS; verifying
  Proposition \ref{prop:post_hoc_wis}. This is not true of PostMinMax, the most
  noticeable example being data set 19 in the top figure.}
\label{fig:iso_prop_verification} 
\end{figure}

Finally, Figure \ref{fig:iso_prop_verification} gives an empirical verification
of Proposition \ref{prop:post_hoc_wis}, which recall, showed that sorting and
PAVA applied in post-processing can never make WIS worse. We can see that for
each of the 34 data sets, combining PostSort or PostPAVA with CrossPenalty (left
panel) or AdaptCrossPenalty (right panel) never hurts WIS, and occasionally
improves it. This is \emph{not} true for PostMinMax, as we can see that
PostMinMax hurts WIS in a few data sets across the panels, the most noticeable
instance being data set 19 on the left.

\subsection{Evaluating conformal calibration}

\begin{figure}[p]
\includegraphics[width=0.99\textwidth]{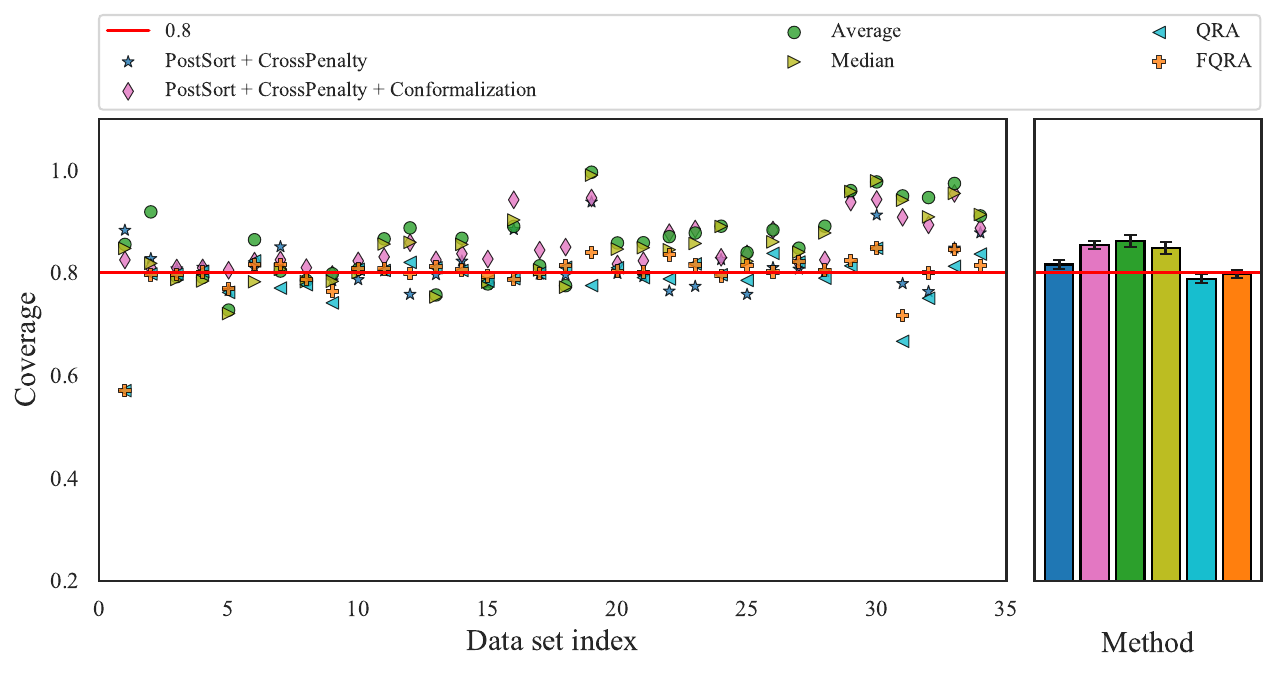}
\captionsetup{format=hang}
\caption{\it\small Comparing coverage of DQA, conformalized DQA, and other
  aggregators at the nominal 0.8 level. The display is as in Figure
  \ref{fig:ensembling_strategies} (individual results per data set on left
  panel, averaged results on right panel). The key point is that conformalized
  DQA manages to achieve valid coverage on each individual data set.}
\label{fig:coverage}

\bigskip\bigskip
\includegraphics[width=0.99\textwidth]{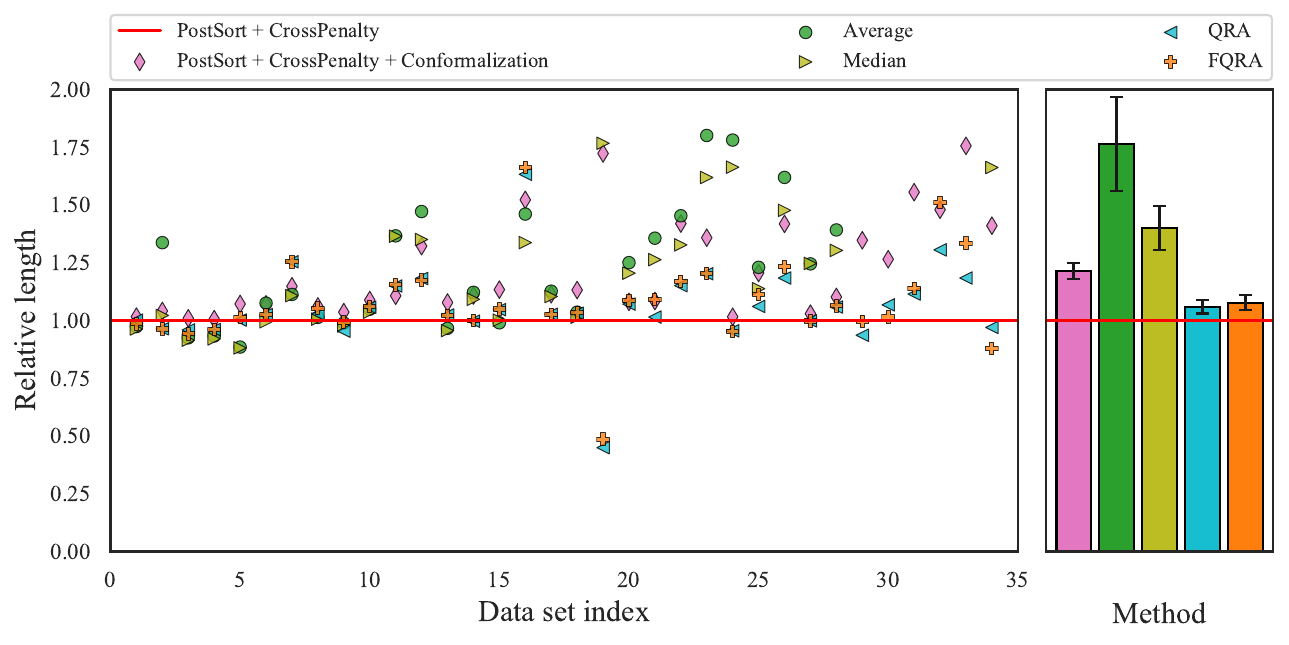}
\captionsetup{format=hang}
\caption{\it\small Comparing prediction interval lengths from DQA, conformalized
  DQA, and others at the nominal 0.8 level. The display is again as in Figure  
  \ref{fig:ensembling_strategies} (individual results on the left, averaged
  results on the right). Conformalized DQA inflates interval lengths compared to
  DQA (in order to achieve valid coverage), but not by a huge amount, making it 
  still clearly favorable to the Average and Median aggregators, and comparable
  to QRA.}  
\label{fig:length}
\end{figure}

We examine the CQR-CV+ method from Section \ref{sec:conformal} applied to DQA,
to adjust its central prediction intervals to have proper coverage. Figures
\ref{fig:coverage} and \ref{fig:length} summarize empirical coverage and
interval length at the nominal 0.8 coverage level, respectively. Note that the results
for other nominal levels are similar. Here, the coverage and length of a
quantile regressor \smash{$\hat{g}$} at level $1-\alpha$ are defined as      
\begin{gather*}
\frac{1}{m} \sum_{i=1}^m 1 \Big\{ \hat{g}(X^*_i; \alpha/2) \leq Y^*_i \leq 
\hat{g}(x_i; 1-\alpha/2) \Big\}, \\
\frac{1}{m} \sum_i^m \Big[ \hat{g}(X^*_i; 1-\alpha/2) - \hat{g}(X^*_i; \alpha/2)
\Big],
\end{gather*}
respectively, where $(X^*_i, Y^*_i)$, $i=1,\ldots,m$ denotes the test set.  The
figures show DQA (denoted CrossPenalty + PostSort), its conformalized version
based on \eqref{eq:cqr_cv+} (denoted CrossPenalty + PostSort +
Conformalization), and the Average, Median, QRA, and FQRA aggregators. 

We can see in Figure \ref{fig:coverage} that DQA achieves fairly close to the
desired 0.8 coverage when this is averaged over all data sets, but it fails to
achieve this \emph{per data set}. Crucially, its conformalized version achieves
at least 0.8 coverage on \emph{every individual data set}. Generally, the
Average and Median aggregators tend to overcover, but there are still data sets
where they undercover. QRA and FQRA have fairly accurate average coverage over
all data sets, but also fail to cover on several individual data sets.

As for length, we can see in Figure \ref{fig:length} that conformalizing DQA
often inflates the intervals, but never by more than 75\% (a relative factor of
1.75), and typically only in the 0-30\% range. The Average and Median
aggregators, particularly the former, tend to have longer intervals by
comparison. QRA and FQRA tend to output slightly longer intervals than DQA, and
slightly shorter intervals than conformalized DQA.

It is worth emphasizing that conformalized DQA is the only method among those
discussed here that is completely robust to the calibration properties of the
constituent quantile regression models (as is true of conformal methods in
general). The coverage exhibited by the other aggregators will generally be a
function of that of the base models, to varying degrees (e.g., the Median
aggregator is more robust than the Average aggregator).

\section{Discussion}
\label{sec:discussion} 

We studied methods for aggregating any number of conditional quantile models
with multiple quantile levels, introducing weighted ensemble schemes with
varying degrees of flexibility, and allowing for weight functions that vary over 
the input feature space. Being based on neural network architectures, our
quantile aggregators are easy and efficient to implement in deep learning
toolkits. To impose proper noncrossing of estimated quantiles, we examined
penalties and isotonization operators that can be used either post hoc or during
training. To guarantee proper coverage of central prediction intervals, we gave
an efficient nested application of the conformal prediction method CQR-CV+
within our aggregation framework.  

Finally, we carried out a large empirical comparison of all of our proposals and
others from the literature across 34 data sets from two popular benchmark
repositories, the most comprehensive evaluation quantile regression/aggregation
methods that we know of. The conclusion is roughly that the most flexible model
in our aggregation family, which we call DQA (deep quantile aggregation),
combined with the simplest penalty and simplest post processor (just post
sorting the quantile predictions), and conformalized using CQR-CV+, is often 
among the most accurate and well-calibrated aggregation methods available, and
should be a useful method in the modern data scientist's toolbox.

{\RaggedRight
\bibliography{quantile}
}

\clearpage
\appendix

\section{Proofs}

\subsection{Proof of equivalence \eqref{eq:crps_pinball}}
\label{app:crps_pinball}

At the outset, we assume two conditions on $F$: it has a derivative $f$, and it
yields an expectation. Starting with the right-hand side in
\eqref{eq:crps_pinball}, we can use substitute $Y=F^{-1}(\tau)$ to rewrite the
integral as  
\[
2 \int_0^1 (1\{Y \leq F^{-1}(\tau)\} - \tau) (F^{-1}(\tau)-Y) \, \d\tau =  
2 \int_{-\infty}^\infty (1\{Y \leq y\} - F(y)) (y-Y) f(y) \, \d{y}.
\]
Let $u'(y)=2(1\{Y \leq y\} - F(y)) f(y)$ and $v(y)=y-Y$. The idea is now to use
integration by parts, but there are two subtleties. First, one has to be careful
about framing the antiderivative $u$ of $u'$, since $y \mapsto 1\{Y \leq y\}$ is
not classically differentiable. Note that we can actually redefine $u'$ to be   
\[
u'(y) = 2(1\{Y \leq y\} - F(y)) (f(y) - \delta_Y(y)),
\]
where $\delta_Y$ is the Dirac delta function centered at $Y$, because the
``extra'' piece integrates to zero:   
\[
\int_{-\infty}^\infty 2(1\{Y \leq y\} - F(y)) (y-Y) \delta_Y(y) \, \d{y} =  
2(1\{Y \leq y\} - F(y)) (y-Y) \Big|_{y=Y} = 0. 
\]
With this new definition of $u'$, its antiderivative is rigorously
\[
u(y) = -(1\{Y \leq y\} - F(y))^2,
\] 
because, in the distributional sense, the derivative of the heavyside function
$y \mapsto 1\{Y \leq y\}$ is indeed the delta function $\delta_Y$. Thus we have   
\begin{align*}
\int_{-\infty}^\infty u'(y) v(y) \, \d{y} 
&= u(y) v(y) \Big|_{-\infty}^\infty - 
\int_{-\infty}^\infty u(y) v'(y) \,  \d{y} \\ 
&= -(1\{Y \leq y\} - F(y))^2 (y-Y) \Big|_{-\infty}^\infty + 
\int_{-\infty}^\infty (1\{Y \leq y\} - F(y))^2 \, \d{y}. 
\end{align*}
The second subtlety is to show that the first term above is indeed zero.  
This is really a question of how fast the tails of $F$ decay. As $F$ yields an
expectation, note that we must have $1-F(y) \lesssim y^{-p}$ for $p>1$ (since
$1/y$ is not integrable on any one-sided interval $[q,\infty)$ for $q>0$). Hence   
\[
(1-F(y)) y \lesssim y^{-p+1} \to 1 \quad \text{as $y \to \infty$},
\]
and the other limit, as $y \to -\infty$, is similar.

\subsection{Proof of equivalence \eqref{eq:wis_pinball}}
\label{app:wis_pinball}

Starting with the right-hand side in \eqref{eq:wis_pinball}, consider the two
summands corresponding to $\tau \in \{\alpha/2, 1-\alpha/2\}$:
\begin{equation}
\label{eq:pinball_summands}
2\pinball_{\alpha/2}(Y - \ell_\alpha) + 2\pinball_{\alpha/2}(Y - u_\alpha),
\end{equation}
where we have used $\ell_\alpha= F^{-1}(\alpha/2)$ and $u_\alpha =
F^{-1}(1-\alpha/2)$. If $Y > u_\alpha$, then \eqref{eq:pinball_summands} is 
\[
\alpha (Y - \ell_\alpha) + 2(1-\alpha/2) (Y - u_\alpha) = 
\alpha (u_\alpha - \ell_\alpha) + 2(Y - u_\alpha).
\]
Meanwhile, if $Y < \ell_\alpha$, then \eqref{eq:pinball_summands} is 
\[
2(1-\alpha/2) (\ell_\alpha - Y) + \alpha (u_\alpha - Y) = 
\alpha (u_\alpha - \ell_\alpha) + 2(\ell_\alpha - Y).
\]
Lastly, if $Y \in [\ell_\alpha, u_\alpha]$, then \eqref{eq:pinball_summands} is 
\[
\alpha (Y - \ell_\alpha) + \alpha (u_\alpha - Y) = \alpha (u_\alpha -
\ell_\alpha).
\]
In each case, it matches the summand in the left-hand side of
\eqref{eq:wis_pinball} corresponding to exclusion probability $\alpha$, which
completes this proof. 

\subsection{Proof of Proposition \ref{prop:post_hoc_wis}}
\label{app:post_hoc_wis}

In this proof we denote the pinball loss, for a vector of quantiles $q \in \R^m$  
across quantiles levels $\tau_1, \ldots, \tau_m$, by
\[
L(q, y) = \sum_{i=1}^m \pinball_{\tau_i} (y - q_i).
\] 

\subsubsection{Proof of part (i)} 

Suppose that there are only two quantile levels $\tau_1 < \tau_2$, and
abbreviate \smash{$q_i = \hat{g}(x; \tau_i)$} for $i=1,2$. We will prove in this 
special case that sorting can only improve the pinball loss. Importantly, this
suffices to prove the result in the general case as well, since sorting can
always be achieved by a sequence of pairwise swaps (via bubble sort).

Assume $q_1 > q_2$, otherwise there is nothing to prove. Denote the pinball loss
by 
\[
L(q,y) = \pinball_{\tau_1} (y - q_1) + \pinball_{\tau_2} (y - q_2) .
\]
We now break the argument down into three cases. In each case, when we speak of
the loss difference, we mean the pre-sort minus the post-sort pinball loss
(first line minus second line in each display that follows).   

\medskip\noindent
\underline{Case 1:} $y > q_1$. Denoting $a = y-q_1$ and $b = y-q_2$ (and noting
$b \geq a$), we have     
\begin{align*}
L((q_1,q_2), y) &= \tau_1 a + \tau_2 b, \\
L((q_2,q_1), y) &= \tau_1 b + \tau_2 a, 
\end{align*}
so the loss difference is $(\tau_2 - \tau_1)(b-a) \geq 0$. I

\medskip\noindent
\underline{Case 2:} $y \in [q_2,q_1]$. Denoting $a = q_1-y$ and $b=y-q_2$, we
have 
\begin{align*}
L((q_1,q_2), y) &= (1-\tau_1) a + \tau_2 b, \\
L((q_2,q_1), y) &= \tau_1 b + (1-\tau_2) a, 
\end{align*}
so the loss difference is $(\tau_2 - \tau_1)(a+b) \geq 0$. 

\medskip\noindent
\underline{Case 3:} $y < q_2$. Denoting $a = q_1-y$ and $b = q_2-y$ (and noting
$a \geq b$), we have 
\begin{align*}
L((q_1,q_2), y) &= (1-\tau_1) a + (1-\tau_2) b, \\
L((q_2,q_1), y) &= (1-\tau_1) b + (1-\tau_2) a, 
\end{align*}
so the loss difference is $(\tau_2 - \tau_1)(a-b) \geq 0$. This completes the 
proof. 

\subsubsection{Proof of part (ii)}

As in the proof of part (i), we will prove the result in a special case, and
argue that it suffices to prove the result in general. Given $k+\ell$ quantile
levels $\tau_1 < \cdots < \tau_{k+\ell}$, we abbreviate \smash{$q_i = \hat{g}(x;
\tau_i)$} for $i=1,\ldots,k+\ell$. We suppose, in what follows, that $q_1 =
\cdots = q_k$ and $q_{k+1} = \cdots = q_{k+\ell}$. We then show that 
averaging can only improve the pinball loss, where by averaging, we mean
replacing the original vector $(q_1,\ldots,q_{k+\ell})$ by \smash{$(\bar{q},
\ldots, \bar{q})$}, with \smash{$\bar{q} = \frac{1}{k+\ell} \sum_{i=1}^{k+\ell}
q_i$}. This suffices to prove the desired result, because isotonic projection 
can be solved using PAVA, which carries out a sequence of steps (whenever there
is a violation of the monotonicity condition) that are precisely of the form we
study here. 

Assume $q_k > q_{k+1}$, otherwise there is nothing to prove. We break the
argument down into four cases. In each case, as in the proof of part (i), when 
we speak of the loss difference below, we mean the pre-sort loss minus the
post-sort loss.   

\medskip\noindent
\underline{Case 1:} $y \geq q_k$. We have  
\begin{align*}
L((q_1, \ldots, q_{k+\ell}), y) &= \sum_{i=1}^k \tau_i (y - q_k) + 
\sum_{i=k+1}^{k+\ell} \tau_i (y - q_{k+1}), \\
L((\bar{q}, \ldots, \bar{q}), y) &= \sum_{i=1}^{k+\ell} \tau_i (y - \bar{q}),
\end{align*}
so the loss difference is 
\[
\sum_{i=1}^k \tau_i (\bar{q} - q_k) + 
\sum_{i=k+1}^{k+\ell} \tau_i (\bar{q} - q_{k+1}) 
= \frac{k\ell}{k+\ell} 
\bigg( \frac{1}{\ell} \sum_{i=k+1}^{k+\ell} \tau_i - 
\frac{1}{k} \sum_{i=1}^k \tau_i \bigg)(q_k - q_{k+1}) \geq 0.
\]

\medskip\noindent
\underline{Case 2:} $y \in [q_{k+1}, q_k]$ and \smash{$y > \bar{q}$}. We have   
\begin{align*}
L((q_1, \ldots, q_{k+\ell}), y) &= \sum_{i=1}^k (1-\tau_i) (q_k - y) +  
\sum_{i=k+1}^{k+\ell} \tau_i (y - q_{k+1}), \\
L((\bar{q}, \ldots, \bar{q}), y) &= \sum_{i=1}^{k+\ell} \tau_i (y - \bar{q}),
\end{align*}
so the loss difference is 
\[
k (q_k - y) + \sum_{i=1}^k \tau_i (\bar{q} - q_k) + 
\sum_{i=k+1}^{k+\ell} \tau_i (\bar{q} - q_{k+1}) 
= k (q_k - y) + \frac{k\ell}{k+\ell} 
\bigg( \frac{1}{\ell} \sum_{i=k+1}^{k+\ell} \tau_i - 
\frac{1}{k} \sum_{i=1}^k \tau_i \bigg)(q_k - q_{k+1}) \geq 0.
\]

\medskip\noindent
\underline{Case 3:} $y \in [q_{k+1}, q_k]$ and \smash{$y \leq \bar{q}$}. We have    
\begin{align*}
L((q_1, \ldots, q_{k+\ell}), y) &= \sum_{i=1}^k (1-\tau_i) (q_k - y) +  
\sum_{i=k+1}^{k+\ell} \tau_i (y - q_{k+1}), \\
L((\bar{q}, \ldots, \bar{q}), y) &= \sum_{i=1}^{k+\ell} (1-\tau_i) (\bar{q} -y), 
\end{align*}
so the loss difference is 
\begin{multline*}
\ell (y - q_{k+1}) + \sum_{i=1}^k(1-\tau_i) (q_k - \bar{q}) + 
\sum_{i=k}^{k+\ell}(1-\tau_i) (q_{k+1} - \bar{q}) \\
= \ell (y - q_{k+1}) + \frac{k\ell}{k+\ell} 
\bigg( \frac{1}{k} \sum_{i=1}^k (1-\tau_i) - 
\frac{1}{\ell} \sum_{i={k+1}}^{k+\ell} (1-\tau_i) \bigg) (q_k - q_{k+1}) \geq 0.
\end{multline*}

\medskip\noindent
\underline{Case 4:} $y < q_k$. Then 
\begin{align*}
L((q_1, \ldots, q_{k+\ell}), y) &= \sum_{i=1}^k (1-\tau_i) (q_k - y) +  
\sum_{i=k+1}^{k+\ell} (1-\tau_i) (q_{k+1} - y), \\
L((\bar{q}, \ldots, \bar{q}), y) &= \sum_{i=1}^{k+\ell} (1-\tau_i) (\bar{q} -y), 
\end{align*}
so the loss difference is 
\[
\sum_{i=1}^k(1-\tau_i) (q_k - \bar{q}) + 
\sum_{i=k}^{k+\ell}(1-\tau_i) (q_{k+1} - \bar{q}) 
= \frac{k\ell}{k+\ell} 
\bigg( \frac{1}{k} \sum_{i=1}^k (1-\tau_i) - 
\frac{1}{\ell} \sum_{i={k+1}}^{k+\ell} (1-\tau_i) \bigg) (q_k - q_{k+1}) \geq 0.
\]
This completes the proof. 

\subsection{Proof of Proposition \ref{prop:tail_behavior}}
\label{app:tail_behavior}

\subsubsection{Proof of part (i)}

This is immediate from the fact that $f(v) = w_1 f_1(v) + w_2 f_2(v)$ and
$f_2(v) / f_1(v) \to 0$.   

\subsubsection{Proof of part (ii)}

This part is more subtle. Using \eqref{eq:probability_quantile}, note that we
may write  
\[
\frac{1}{\bar{f}(\bar{Q}(u))} = \frac{w_1}{f_1(Q_1(u))} +
\frac{w_2}{f_2(Q_2(u))}. 
\]
For \smash{$v = \bar{Q}(u)$}, consider 
\[
\frac{f_1(v)}{\bar{f}(v)} = \frac{w_1f_1(\bar{Q}(u))}{f_1(Q_1(u))} +
\frac{w_2f_1(\bar{Q}(u))}{f_2(Q_2(u))}. 
\] 
It will be convenient to work on the log scale.  Introduce $p_1 = \log f_1$ and
$p_2 = \log f_2$.  Then it suffices to show that as $u \to 1$,
\begin{gather*}
p_1(\bar{Q}(u)) - p_1(Q_1(u)) \to \infty, \quad \text{and} \\ 
p_1(\bar{Q}(u)) - p_2(Q_2(u)) \; \text{remains bounded away from $-\infty$}, 
\end{gather*}
or
\begin{gather*}
p_1(\bar{Q}(u)) - p_2(Q_2(u)) \to \infty, \quad \text{and} \\ 
p_1(\bar{Q}(u)) - p_1(Q_1(u)) \; \text{remains bounded away from $-\infty$}. 
\end{gather*}
We now divide the argument into two cases. Without a loss of generality, we set
$w_1=w_2=1/2$.  

\medskip\noindent
\underline{Case 1:} $p_1(Q_1(u)) - p_2(Q_2(u))$ remains bounded away from 
$-\infty$. Denoting \smash{$v=\bar{Q}(u)$}, $v_1=Q_1(u)$, $v_2=Q_2(u)$, we have,
using log-concavity of $p_1$,   
\begin{align*}
p_1(v) - p_2(v_2) &= p_1(v_1/2 + v_2/2) - p_2(v_2) \\
&\geq p_1(v_1)/2 + p_1(v_2)/2 - p_2(v_2) \\
&= \big(p_1(v_1) - p_2(v_2)\big)/2 + \big(p_1(v_2) - p_2(v_2)\big)/2.
\end{align*}
The first term above is bounded below by assumption; the second term diverges to
$\infty$ (under our hypothesis that $f_2(v) / f_1(v) \to 0$).  

\medskip\noindent
\underline{Case 2:} $p_1(Q_1(u)) - p_2(Q_2(u)) \to -\infty$. We have, just as in
the first case,  
\begin{align*}
p_1(v) - p_1(v_1) &= p_1(v_1/2 + v_2/2) - p_1(v_1) \\
&\geq p_1(v_1)/2 + p_1(v_2)/2 - p_2(v_2) + p_2(v_2) - p_1(v_1) \\
&= -\big(p_1(v_1) - p_2(v_2)\big)/2 + \big(p_1(v_2) - p_2(v_2)\big)/2.
\end{align*}
Now both terms above diverge to $\infty$, and this completes the proof. 

\section{More experimental details and results}

In the following, we provide further details about models: the hyperparameter
space, optimization details, etc. In our experiments, for each method  
described below, we perform hyperparameter tuning over a randomly chosen subset 
of 20 values from the full Cartesian product of possibilities.

\subsection{Base models}
\label{app:base_models}

We consider $p=6$ base models in total. The first 3 are neural network models,
that each compute quantile estimates differently. Each use a standard
feed-forward (multilayer perceptron) architecture. 

\begin{itemize}  
\item Conditional Gaussian network (CGN,
  \citealp{lakshminarayanan2017uncertainty}):   
  this model assumes that the conditional distribution of $Y|X=x$ is
  Gaussian. It therefore outputs estimates of sufficient statistics, namely the
  conditional mean $\mu(x)$ and variance $\sigma^2(x)$. It is optimized by
  minimizing the negative log-likelihood, and subsequent quantile estimation 
  is done based on the normal distribution.
  
\item  Simultaneous quantile regression (SQR, \citealp{tagasovska2019single}):
  this model takes as input a target quantile level $\tau$ concatenated with
  a feature vector $x$, and outputs a corresponding quantile estimate. It is
  trained by minimizing a pinball loss for many different randomly sampled
  quantile levels $\tau$. To estimate multiple quantile levels with SQR, we must
  thus feed in the same $x$ numerous times, each paired with a different
  quantile level $\tau$.  

\item Deep quantile regression (DQR): this model falls naturally out of our
  neural aggregation framework, as explained in Section \ref{sec:local_aggr}. It
  is the only neural network based model considered in this paper that can
  easily utilize different isotonization approaches discussed in Section
  \ref{sec:noncrossing}. For DQR as a base model in all experiments, we use the
  simplest method for ensuring noncrossing among the all options: the crossing
  penalty along with post sorting. 
\end{itemize}

We optimize each of these neural network models using Adam
\citep{kingma2015adam}, and using ELU activation function
\citep{clevert2016fast}. We adaptively vary the mini-batch size depending on 
the data set size. %(i.e., $2^{3 + \floor{\log_{10}{n}}}$). 
They also share the same architecture/optimization hyperparameter search
space: \# of fully connected layers: \{2, 3\}, \# of hidden units: \{64, 128\},
dropout ratio: \{0.0, 0.05, 0.1\}, learning rate: \{1e-3, 3e-4\}, weight decay:
\{1e-5, 1e-7\}. In all settings, we use early stopping where the validation loss 
is evaluated every epoch and if it has not decreased for the last 500 updates, 
the optimization is stopped by returning the epoch with the lowest validation
loss. 

The next 3 base models are not neural networks.

\begin{itemize}
\item Quantile random forest (RandomForest, \citealp{meinshausen2006quantile})
  and extremely randomized trees (ExtraTrees, \citealp{geurts2006extremely}):
  both models are based on random forests trained as regular (conditional
  mean) regressors. After training, quantile estimates are computed via
  empirical quantiles of the data in relevant leaf nodes of the trees.  

\item Quantile gradient boosting (LightGBM, \citealp{ke2017gbm}): this model
  is a gradient boosting framework that uses tree-based base learners. As a 
  boosting framework, it can optimize the pinball loss, and thus be directly
  used for quantile regression. It handles multiple quantile levels by simply
  using a separate model for each level.
\end{itemize}

For the random forests models, we use the {\tt scikit-garden}
(\url{https://scikit-garden.github.io/}) implementation for both, and both have the 
  same hyperparameter search space: minimum \# of samples for splitting nodes:
  \{8, 16, 64\}, minimum \# of sample for leaf nodes: \{8, 16, 64\} . 
 
For the gradient boosting model, the  hyperparameter space is: \# of leaves:
\{10, 50, 100\}, minimum child samples: \{3, 9, 15\}, minimum child weight:
\{1e-2, 1e-1, 1\}, subsample ratio: \{0.4, 0.6, 0.8\}, subsample ratio of
columns: \{0.4, 0.6\}, $\ell_1$ regularization weight: \{1e-1, 1, 5\}, $\ell_2$
regularization weight: \{1e-1, 1, 5\}.  

\subsection{Aggregation models}
\label{app:aggr_models}

We consider 4 aggregation models from outside framework: Average, Median, 
quantile regression averaging (QRA, \citealp{nowotarski2015computing}), and
factor QRA (FQRA, \citealp{maciejowska2016probabilistic}). The first 3 do not
require tuning. For FQRA, we tune over the number of factors, between 1-6 (the
number of base models).   

We also study a total of 6 aggregation models within our framework: three
global aggregators and three local ones (each having coarse, medium, or fine
weight parameterizations).  For the global aggregators, the hyperparameter
search space 
we use is: crossing penalty weight $\lambda$: \{0.5, 1.0, 2.0, 5.0, 10.0\},
crossing penalty margin scaling $\delta_0$: \{1e-1, 5e-2, 1e-2, 1e-3, 1e-4\}.  
For the local aggregators, the hyperparameter search space additionally
includes: \# of layers: \{2, 3\}, \# of hidden units: \{64, 128\}, dropout
ratio: \{0.0, 0.05, 0.1\}. All are optimized using the Adam
optimizer~\citep{kingma2015adam}, ELU activation function
\citep{clevert2016fast}, and adaptively-varied mini-batch size. We also employ
the same early stopping strategy as described above.  

\subsection{Revisiting Figure \ref{fig:intro}}
\label{app:intro_revisit}

\begin{figure}[t]
\includegraphics[width=\textwidth]{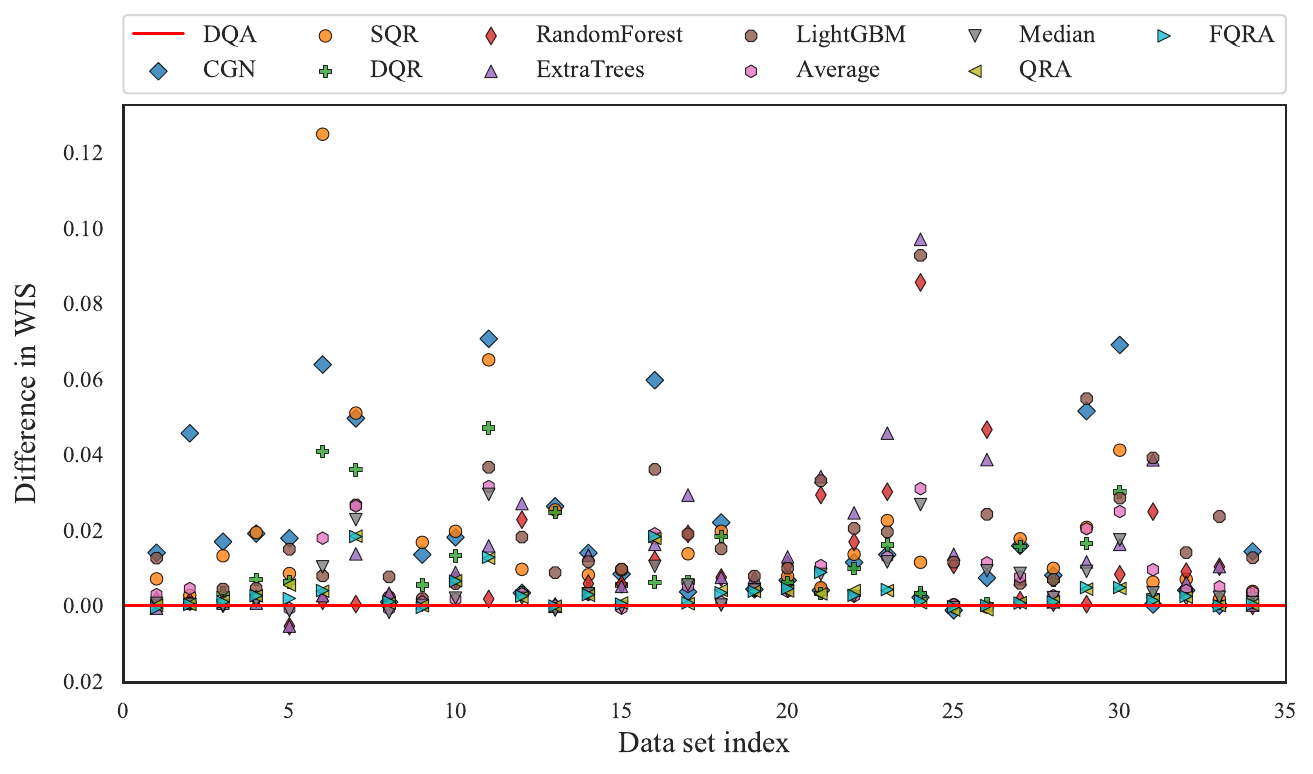}
\captionsetup{format=hang}
\caption{\it\small As in Figure \ref{fig:intro}, but with WIS differences (of
  each method to DQA) on the y-axis rather than WIS ratios.} 
\label{fig:intro_revisit}
\end{figure}

Figure \ref{fig:intro_revisit} plots the same results in Figure \ref{fig:intro}
but it displays a difference in WIS values, rather than a ratio of WIS values
(which is how we define relative WIS). That is, shown on the y-axis is the
difference of the WIS of each quantile regression minus that of DQA, where now 0
indicates equal WIS performance to DQA, and a number greater than 0 indicates
worse performance than DQA. In general, WIS will be on the scale of the response
variable, but recall that we have standardized the response variable in each
data set, thus the comparison in Figure \ref{fig:intro_revisit} makes sense
across data sets. The story in this figure is very much similar to that in
Figure \ref{fig:intro}, except that we can more clearly see that for some data 
sets with low PVE values (e.g., at index 5), there are a handful of models that
perform a little better than DQA.

\end{document}